%% file: main.tex
\documentclass[]{fairmeta}
\input{preamble}

\title{Unified Text-Image Generation with Weakness-Targeted Post-Training}

\author[1,2*]{Jiahui Chen}
\author[1,2]{Philippe Hansen-Estruch}
\author[1]{Xiaochuang Han}
\author[1]{Yushi Hu}
\author[1]{Emily Dinan}
\author[1,3,4]{Amita Kamath}
\author[1]{Michal Drozdzal}
\author[1]{
Reyhane Askari-Hemmat}
\author[1]{
Luke Zettlemoyer}
\author[1]{
Marjan Ghazvininejad}

\affiliation[1]{FAIR at Meta SuperIntelligence Labs}
\affiliation[2]{University of Texas, Austin}
\affiliation[3]{University of Washington}
\affiliation[4]{University of California, Los Angeles}

\contribution[*]{Work done at Meta}
\abstract{\input{sec/abstract}}

\date{\today}
\correspondence{Jiahui Chen \email{jiahui.k.chen@utexas.edu}, Marjan Ghazvininejad \email{ghazvini@meta.com}}

\begin{document}

\maketitle

\vspace{-0.2cm}
\input{sec/intro}
\input{sec/related_work}

\input{sec/rwr}
\input{sec/weaknesses_and_dataset}

\input{sec/rewards}

% % Only one section discussing all results (discusses text+image worse than image-only)
% \input{sec/single_sec_modality_experiments}
% One subsection focused on text+image results, then SEPARATE subsection on when text+image is worse
\input{sec/modality_experiments}

\input{sec/multimodal_failures}
\input{sec/dataset_experiments}

\input{sec/conclusion}

\clearpage
\newpage
\bibliographystyle{plainnat}
\bibliography{main}

\clearpage
\newpage
\beginappendix
\setcounter{page}{1}
\input{sec/X_suppl}

\end{document}

%% file: preamble.tex
\usepackage{adjustbox}
\usepackage{subcaption}
\usepackage{multirow} 
\usepackage{xcolor}         % colors
\usepackage{booktabs}       % professional-quality tables
\usepackage{colortbl}
\usepackage{amsmath}
\usepackage{microtype}
\usepackage{makecell}
\usepackage{enumitem}
\usepackage{float}
\usepackage[most]{tcolorbox}
\tcbuselibrary{listings} 
\tcbset{
  promptbox/.style={
    colback=white,
    colframe=metablue,
    boxrule=0.5pt,
    arc=2pt,
    left=4pt,right=4pt,top=4pt,bottom=4pt,
    listing only,
    listing only,
    listing options={
      basicstyle=\ttfamily\normalsize, % typewriter, same size as main text
      breaklines=true,
      columns=fullflexible,
      keepspaces=true
    }
  }
}

\newcommand{\vstarttok}{\texttt{\textless|vision\_start|\textgreater}}
\newcommand{\thinkendtok}{\texttt{\textless/think\textgreater}}

\definecolor{best_result_color}{RGB}{198,239,206}      % Light green
\definecolor{second_best_result_color}{RGB}{255,229,204}    % Light peach
\definecolor{best_multimodal_border}{RGB}{65,105,225}  % Royal blue
\definecolor{best_imageonly_border}{RGB}{138,43,226}   % Blue violet

\newcommand{\multimodalbox}[1]{{\setlength{\fboxrule}{1.5pt}\fcolorbox{best_multimodal_border}{white}{#1}}}
\newcommand{\imageonlybox}[1]{{\setlength{\fboxrule}{1.5pt}\fcolorbox{best_imageonly_border}{white}{#1}}}
\newcommand{\multimodalboxbest}[1]{{\setlength{\fboxrule}{1.5pt}\fcolorbox{best_multimodal_border}{best_result_color}{#1}}}  % Multimodal Best is OVERALL BEST
\newcommand{\imageonlyboxbest}[1]{{\setlength{\fboxrule}{1.5pt}\fcolorbox{best_imageonly_border}{best_result_color}{#1}}}  % Image-Only Best is OVERALL BEST
\newcommand{\imageonlyboxsecondbest}[1]{{\setlength{\fboxrule}{1.5pt}\fcolorbox{best_imageonly_border}{second_best_result_color}{#1}}}  % Image-Only Best is OVERALL BEST

%% file: sec/abstract.tex
% No abstract environment here: fairmeta.cls redefines \abstract as a one-argument
% command (main.tex calls \abstract{\input{sec/abstract}}), so \begin{abstract}
% would swallow the first character of the text.
Unified multimodal generation architectures that jointly produce text and images have recently emerged as a promising direction for text-to-image (T2I) synthesis. 
In one approach, the model first generates a textual reasoning trace, and a manual modality switch in the inference pipeline then starts image generation.
We study reward-weighted post-training with fully self-generated synthetic data for this reasoning-conditioned generation mode on BAGEL, a 14B mixture-of-transformers model that pairs autoregressive text generation with flow-matching image synthesis. Our post-training enables a learned modality switch, so the model itself can trigger image generation after its reasoning. We train on prompts that target the model's generation weaknesses, which yield better results than general image-caption or benchmark-derived prompts, a gain we do not attribute to weakness targeting alone.
Our approach improves multimodal image generation on object-centric prompt alignment and text rendering, while remaining comparable on dense-prompt alignment and knowledge-intensive generation.
We examine the impact of joint text-image generation on T2I performance and the relative importance of each modality during post-training. We apply the reward weight to the text loss, the image loss, or both, and compare against image-only post-training on the same images and reward labels. These modality ablations show that weighting both losses is most effective, and that reward weighting outperforms supervised fine-tuning in both multimodal and image-only post-training, especially for text rendering. A comparison of the two generation modes shows that reasoning helps on knowledge-intensive prompts, while image-only generation remains stronger on dense-prompt alignment and text rendering.

%% file: sec/intro.tex
\section{Introduction}
\vspace{-0.2cm}
Text-to-image (T2I) generation has evolved significantly with the advent of multimodal generative models that output both text and images within a unified architecture~\cite{show-o,transfusion, chameleon}. 
As the field moves toward omni-modal systems that handle multiple modalities simultaneously, a new class of multimodal T2I models has emerged that utilize intermediate text generation to improve image synthesis. 
Recent multimodal architectures such as BAGEL~\cite{bagel} exemplify this approach by employing a two-stage inference process: first generating text as an intermediate ``reasoning'' step, then switching to image generation conditioned on the produced reasoning text. 
This approach posits that explicit textual reasoning can guide and improve the subsequent image generation process, enabling models to better capture complex semantic relationships and nuanced visual details. 
In practice, however, the intermediate reasoning does not reliably help: on several benchmarks BAGEL's reasoning-conditioned (multimodal) generation underperforms its own image-only generation~\cite{oneigbenchomnidimensionalnuancedevaluation, dpgbench, genaibenchevaluatingimprovingcompositional}; see Tables \ref{tab:dpgbench_modality} and \ref{tab:oneigbench_modality}.
In addition, the switch from reasoning to image generation is typically controlled by the inference pipeline (a manual modality switch), for example by stopping text decoding at a delimiter and launching a separate image-generation call.

In this work, we study whether post-training on the model's own reasoning-and-image samples can improve reasoning-conditioned T2I generation.
This post-training enables a learned modality switch, where the model emits the image-start token \vstarttok{} after generating its reasoning.
Thus, at inference, the model itself can trigger image generation after reasoning, rather than relying on a manual modality switch as BAGEL's standard multimodal generation does.
As part of our multimodal post-training process, we investigate the relative importance of each modality during training, assessing whether reward-weighting both the text and image losses yields greater improvements than reward-weighting only one. We also compare against image-only post-training on the same images and reward labels, which tests whether the gains require intermediate reasoning.

Building on the analysis of modality interplay during post-training, we additionally examine how data composition shapes model performance. 
The prompts used in post-training determine which samples are collected and optimized, making prompt selection a defining factor in model refinement.
Existing prompt sourcing strategies vary widely: standard T2I post-training typically relies on broad web-scale image--caption datasets~\cite{trainingdiffusionmodelsreinforcement, adjointmatchingfinetuningflow, show-o}, while recent online RL methods use benchmark-specific prompts that mirror evaluation sets~\cite{flowgrpotrainingflowmatching}. 
Yet, the optimal strategy for selecting prompts remains uncertain.
To address this, we systematically compare the standard and benchmark-aligned data strategies with our novel weakness-targeted dataset, and find that it performs best among the sets we compare, although our comparison does not isolate weakness targeting from the other ways the prompt sets differ, such as reward variance, category composition, and semantic overlap with the benchmarks.

\begin{figure}[t]
    \centering
    \includegraphics[width=1.0\textwidth]{figs/fig1_.pdf}
     \vspace{-0.1cm}
    \caption{Sample generations before and after our post-training. Our post-training enables successful generations for challenging, previously failed prompts.}
    \label{fig:fig_1}
\end{figure}

Across four benchmarks with diverse evaluation criteria, reward-weighting both modalities improves BAGEL's multimodal image generation on object-centric prompt alignment and text rendering, while remaining comparable to the baseline on dense-prompt alignment and knowledge-intensive generation.  
Through analysis of six candidate reward functions, we find that a VQAScore-style reward~\cite{genaibenchevaluatingimprovingcompositional} agrees best with independent judgments of generation success from an object detector and from human annotators, and we use it in an offline reward-weighted regression method that trains on fully self-generated synthetic data.
We hope our findings contribute to more effective training of multimodal models, and we summarize our main contributions below:
\begin{itemize}[itemsep=0pt,parsep=0pt,topsep=-3pt,partopsep=0pt]
    \item We show that offline reward-weighted post-training on self-generated data improves BAGEL's multimodal generation, most clearly in object-centric prompt alignment and text rendering (a 5-point GenEval gain and a ninefold increase in OneIG-Bench text rendering over the multimodal baseline). Using matched image-only baselines, we show that reward weighting also improves text rendering without reasoning, and that reasoning helps on knowledge-intensive prompts but hurts dense-prompt alignment and text rendering.
    \item We compare post-training prompt sets and find that the weakness-targeted set gives the largest gains, though the sets also differ in reward variance, category composition, and benchmark overlap, so we do not attribute the gains to weakness targeting alone.
\end{itemize}

%% file: sec/related_work.tex
\section{Related Work}

\vspace{-0.2cm}
\subsection{Multimodal Generative Models}
\vspace{-0.1cm}
We define multimodal generative models as systems that process both text and image modalities as inputs and outputs. Existing approaches differ in architecture, influencing their capacity for joint generation and reasoning.
\textbf{Unified Transformer with Diffusion Components:} share a transformer backbone for both modalities, combining autoregressive (AR) text modeling with diffusion-based image generation; examples include BAGEL~\cite{bagel}, Transfusion~\cite{transfusion}, and Show-o~\cite{show-o}. They rely on diffusion~\cite{denoisingdiffusionprobabilisticmodels} or flow matching~\cite{flowmatchinggenerativemodeling, sd3_esser2024scalingrectifiedflowtransformers} rather than full AR token prediction, limiting seamless interleaving without explicit task tokens. \textbf{Fully Autoregressive Models:} generate vision and language tokens within a single AR stream. Janus-Pro~\cite{januspro} and Chameleon~\cite{chameleon} encode images as discrete tokens in the same vocabulary as text, enabling unified multimodal generation, while UniGen~\cite{unigen} adopts a hybrid approach using MaskGIT-style parallel decoding. \textbf{Hybrid Pipelines:} pair an AR language model with a separate diffusion or de-tokenization module for image synthesis; BLIP3-o~\cite{blip3o} first produces image embeddings then decodes them with a diffusion transformer, and SEED-X~\cite{seedx} generates learnable \texttt{<IMG>} embeddings decoded by a visual module. These designs achieve high visual quality but remain non-unified and block end-to-end multimodal gradient flow.

\vspace{-0.2cm}
\subsection{Multimodal Reasoning for T2I Generation}
\vspace{-0.1cm}
A key goal of multimodal models is leveraging textual reasoning to enhance text-to-image (T2I) synthesis, often by generating intermediate text artifacts or performing self-evaluation during inference.  
Some methods use vision--language reasoning for self-correction. For example, UniGen~\cite{unigen} employs a self-verification mechanism where the model generates multiple image candidates, performs VQA-style analyses to assess prompt alignment, and selects the best result through best-of-\textit{n} filtering.  
Other work explores interleaved generation of text and images. Interleaving Reasoning Generation (IRG)~\cite{irg} iteratively refines outputs by switching between textual reasoning and image synthesis. Such reasoning-driven pipelines improve visual quality and semantic fidelity over prompt-only baselines, though fully integrating reasoning into unified multimodal generators remains an open challenge.

%% file: sec/rwr.tex
% \section{Experimental Setup}
\section{Methodology}

Our base model, BAGEL \cite{bagel}, is a 14B-parameter Mixture-of-Transformers (MoT)~\cite{mixtureoftransformerssparsescalablearchitecture} architecture that utilizes flow matching for image generation. Online RL incurs substantial computational cost, as each denoising timestep of image generation requires a full MoT forward pass. 
We therefore adopt reward-weighted regression (RWR) \cite{rwr_original} for post-training, weighting the training loss by the sample reward (Section \ref{sec:rwr}).
Inspired by online RL's use of sampling for exploration, we use text and image samples from the base model as our training data. 
To target post-training where it matters, we identify systematic visual patterns that consistently cause failed generations and then create a weakness-targeted training prompt set; we present our analysis of high-failure generation categories in Section \ref{sec:multimodal_gen_weaknesses_dataset} and demonstrate the effectiveness of training on our dataset in Section \ref{sec:data_strategies_experiments}.
Finally, we evaluate a suite of reward functions for their ability to differentiate successful from failed generations, for the effective reward-labeling of our dataset (Section \ref{sec:rewards}). 
% Finally, we describe our reward-weighted regression training algorithm which we leverage in all our experiments. 

\subsection{Reward-Weighted Regression with a Learned Image-Start Token}\label{sec:rwr}

Each training sequence consists of the prompt, a reasoning trace ending in \thinkendtok{}, the \vstarttok{} token, and the image, so the model learns to emit \vstarttok{} immediately after its reasoning.
We refer to this as the learned modality switch: at inference, generating the \vstarttok{} token triggers image generation conditioned only on the previously generated reasoning. 
Unlike in BAGEL's standard thinking pipeline, the prompt is not re-inserted, so it influences the image only through the reasoning, which was conditioned on the prompt.
Because the token is emitted at the same point where a \thinkendtok{}-triggered modality switch would occur, learning it does not change when the modality switch happens. Its benefit is that the transition is produced by the model and included in the (reward-weighted) training loss rather than hard-coded in the inference code.
On GenEval, the learned modality switch, a \thinkendtok{}-triggered modality switch with the prompt re-inserted, and a manual modality switch using BAGEL's standard thinking pipeline give statistically indistinguishable scores (Appendix~\ref{app:switching}). This comparison uses our only surviving post-trained multimodal checkpoint, a Multimodal RWR variant whose reward additionally includes an OCR score on long-text prompts (hereafter, the surviving OCR-reward checkpoint), not the Multimodal RWR checkpoint reported in the main tables. 
All text and vision tokens are processed as a packed sequence with gradient updates every 50k tokens; full training details are in Appendix \ref{app:training_hyperparams}.
% All images are limited to a resolution of 512x512 and full details on training hyperparameters are presented in Appendix \ref{app:training_hyperparams}. 
% , and following ~\cite{bagel}, at training time images are tokenized into 3 sets of vision tokens: noised VAE latent tokens used in to optimize the image generation loss, VAE latent tokens used to condition subsequent generation of any modality, and SigLIP2~\cite{siglip2multilingualvisionlanguage} tokens used in input images for visual understanding and to improve image generation quality. 

For reward-weighted regression (RWR), we extend BAGEL's original objective by applying an exponentiated reward weight 
\( w_{\text{RWR}}(x_0, c) = \exp(\beta z(x_0, c)) \), with \( z(x_0, c) = (r(x_0, c) - \mu_r)/\sigma_r \),
where \( r(x_0, c) \) is the reward for training sample \( x_0 \) and prompt \( c \), following prior work~\cite{trainingdiffusionmodelsreinforcement, lee2023aligning}. 
\( \beta \) is the temperature controlling the strength of the exponential weighting; we use \( \beta = 5.0 \). The mean \( \mu_r \) and standard deviation \( \sigma_r \) are computed over all training samples (a global z-score), and weights are not clipped; on MMGW with QwenVQAScore the mean weight is 73.9 (maximum 262). The weight multiplies each sample's token losses. The text cross-entropy loss is normalized by the sum of weights, \( \sum_i w_i \ell^{\text{CE}}_i / \sum_i w_i \), whereas the image flow-matching loss is averaged over image tokens, \( \sum_i w_i \ell^{\text{MSE}}_i / N \). Configurations that reward-weight the image loss therefore train with a larger effective image-loss scale than SFT or CE-weighted RWR, bounded by gradient-norm clipping at 1.0. The total loss is \( \mathcal{L}_{\text{CE}} + 0.1\,\mathcal{L}_{\text{MSE}} \), as in BAGEL. 
Although we also tested rejection sampling, i.e., training only on samples exceeding a reward threshold, RWR outperformed it on every benchmark and metric we report; full results are in Appendix~\ref{app:rejection_sampling}.
BAGEL optimizes MSE loss on image tokens and CE loss on text tokens, allowing reward-weighting of either or both objectives to study modality-specific learning effects (Section~\ref{sec:modality_experiments}).
% Rewards are normalized across all samples before weighting. 
We use a single round of sampling, reward labeling, and RWR training, but in general this approach can be repeated for multiple rounds of alternating sampling and training to enable online RL.

%% file: sec/weaknesses_and_dataset.tex
% MMGW categories
\begin{figure}[]
  \centering
  \vspace{-0.1cm}
  \includegraphics[width=0.9\textwidth]{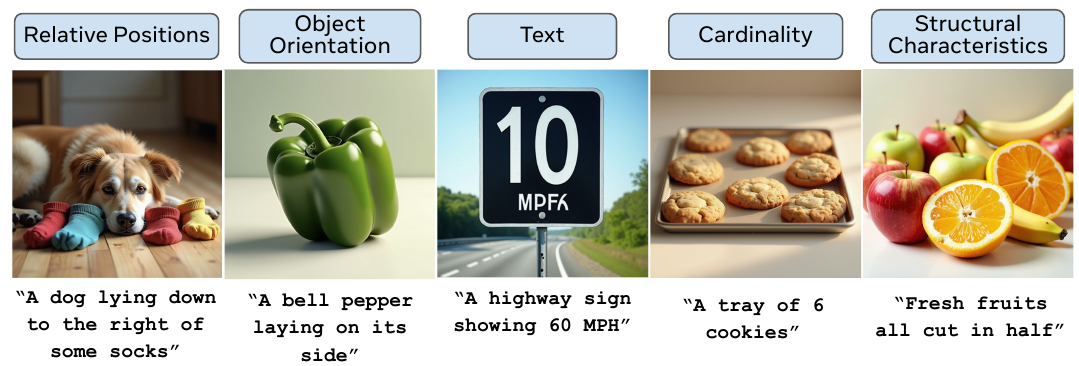}
  % \vspace{-0.2cm}
  \caption{
  The Multimodal Generative Weaknesses (MMGW) Dataset includes five semantic categories that reliably induce generation failures in unified multimodal models. Here we show one representative prompt and failure image for each category.
  % : Relative Positions, Object Orientation, Text, Cardinality, and Structural Characteristics, 
  % Each category originates from challenging visual patterns documented in prior vision-language evaluation work.
  }
  \label{fig:mmgw}
  \vspace{-0.5cm}
\end{figure}

\subsection{Multimodal Generative Weaknesses (MMGW) Dataset}\label{sec:multimodal_gen_weaknesses_dataset}

Our base model, BAGEL~\cite{bagel}, is capable of generating samples with high visual quality, making the selection of effective post-training data non-trivial.
Unlike standard approaches that rely on broad image-caption datasets, we adopt a targeted strategy, identifying prompts that elicit inconsistent generation quality from the base model, where only a subset of sampled outputs are successful. 
The intuition is that isolating prompts with high variance in generation quality, then post-training on many synthetic samples with appropriate reward weighting, will enable the model to learn to consistently produce successful generations; we verify that MMGW prompts exhibit this property in Section~\ref{sec:data_strategies_experiments}. This process could easily be automated by generating prompts and evaluating intra-prompt reward distributions to find prompts producing high-variance reward distributions. 

BAGEL couples a pretrained LLM backbone with a SigLIP2~\cite{siglip2multilingualvisionlanguage} ViT encoder for visual understanding and a separate generation expert that predicts the rectified-flow velocity over frozen VAE latents. Because SigLIP2 tokens also condition multimodal image generation~\cite{bagel}, the model's visual understanding limitations plausibly translate into image generation weaknesses.
We therefore construct our training prompt set from categories known to challenge vision-language models in understanding tasks. The MMVP Benchmark \cite{mmvp_eyeswideshutexploring} identifies 9 visual patterns that vision-language models struggle to interpret. 
We use these patterns along with image captions from MMVP to construct prompts that, through manual testing, result in failed generations approximately 50\% of the time---even when using proprietary multimodal generative models like GPT-4o. 

We find five semantic categories that consistently cause failed generations: Relative Positions, Object Orientation, Text, Cardinality, and Structural Characteristics, with example failures shown in Figure \ref{fig:mmgw}. We call this prompt set the Multimodal Generative Weaknesses (MMGW) Dataset. 
Using approximately 50 manually verified failure-inducing prompts per category (provided in Appendix~\ref{app:mmgw_subset}) as examples, we use Llama-3-70B-Instruct~\cite{llama3modelcard} to generate approximately 500--700 prompts for each of Relative Positions, Object Orientation, Cardinality, and Structural Characteristics, and approximately 1,150 for Text, which spans three sub-types: short and single-sentence rendered text, text within scenes, and text in graphic designs. This yields a final dataset of approximately 3,500 prompts; for each we generate 100 samples of both text reasoning traces and images, for a total of 357,300 training samples. However, to effectively leverage this synthetic data for post-training, we must distinguish which samples represent successful generations versus failures---a challenge we address through reward modeling.

%% file: sec/rewards.tex
\subsection{Effective Reward-Labeling of Self-Generated Synthetic Data}\label{sec:rewards}

We require reward functions capable of distinguishing high-quality from low-quality images, assigning higher rewards to successful generations and lower rewards to failed outputs. Given that our synthetic dataset contains multiple image samples per training prompt, intra-prompt reward variance is essential to enable exploration and differentiate between successful and unsuccessful prompt depictions.

We leverage several existing image reward models: \textsc{PickScore}~\cite{pickapicopendatasetuser}, trained on human-annotated image judgments for producing a score capturing both visual quality and prompt alignment; \textsc{AestheticScore}~\cite{aestheticscore}, using the LAION aesthetics predictor to evaluate image aesthetic quality; \textsc{ImageReward}~\cite{imagerewardlearningevaluatinghuman}, a human preference reward model capturing text-image alignment, visual fidelity, and harmlessness; and \textsc{CLIPScore}~\cite{clipscorereferencefreeevaluationmetric}, measuring CLIP embedding similarity between prompts and images as prompt alignment.
We also formulate two reward functions utilizing existing work. 
\textsc{JPEGScore} is defined as the size of the image after JPEG compression, serving as a measure of incompressibility where higher scores indicate greater complexity~\cite{trainingdiffusionmodelsreinforcement}. 
\textsc{QwenVQAScore} queries Qwen2.5-VL-7B-Instruct~\cite{qwen2.5-VL} with the image, the system prompt ``You are a strict visual judge. Output exactly one word: Yes or No.'', and the question ``Does this image match the following text prompt? Text prompt: $\langle$\textit{p}$\rangle$. Answer with only `Yes' or `No'.'', where $\langle$\textit{p}$\rangle$ is the target image generation prompt, and scores the image as $\sigma\big(\log p(\text{``Yes.''}) - \log p(\text{``No.''})\big)$. This approach adapts VQAScore~\cite{genaibenchevaluatingimprovingcompositional}, originally proposed as a prompt alignment evaluation metric.

Figure~\ref{fig:prompt_reward_dists} shows qualitative examples of per-prompt reward distributions, in which QwenVQAScore effectively labels successful and failed samples, and Figure~\ref{fig:global_reward_dist} (Appendix~\ref{app:reward_dists}) shows the global reward distributions for all six reward functions across all samples in our MMGW dataset.
QwenVQAScore exhibits a bimodal distribution with high-density regions near 0 and 1, while the other rewards are unimodal. Distribution shape alone, however, does not show that a reward separates successful from failed generations, so we validate all rewards against three independent references (Table~\ref{tab:reward_validation}). (i) For all 357,500 BAGEL samples generated from the GenEval-generated prompts (Section~\ref{sec:data_strategies_experiments}), we label each image as correct or incorrect with the GenEval object detector~\cite{geneval}. Applied to base BAGEL's GenEval benchmark images, this labeling reproduces the official GenEval score exactly. (ii) We use human prompt-alignment ratings from GenAI-Bench~\cite{genaibenchevaluatingimprovingcompositional} (9,600 images from six other generators, rated 1--5 by three annotators). (iii) Five annotators, one of them an author, rated 300 of our own MMGW training images (30 prompts drawn at random within categories, 10 images each) as matching, partly matching, or not matching the prompt, blind to the rewards (Fleiss' $\kappa = 0.76$). Because reward-weighted regression exploits differences among samples of the same prompt, our main metric is \emph{within-prompt pairwise accuracy}: among pairs of images of the same prompt that the reference judges differently, the fraction in which the reward ranks the better image higher. For (ii) and (iii), a pair counts as judged differently when its mean ratings differ by at least 1 and 0.5, respectively. We also report AUROC: with images pooled across all prompts, the probability that a randomly chosen good image receives a higher reward than a randomly chosen bad one. Good vs.\ bad is detector-correct vs.\ incorrect for (i), a mean rating of at least 4 vs.\ at most 2 for (ii), and a majority judgment of matching vs.\ not matching for (iii).

% we confirmed that QwenVQAScore can consistently distinguish successful generations from failed ones for the same prompt.
% \yushi{Maybe the claim is too strong here? Qwen is not 100\% correct. Weaken it a bit?} 
Across all three references, QwenVQAScore achieves the highest within-prompt pairwise accuracy (about 0.75 in each) and the highest AUROC. By comparison, ImageReward and PickScore reach 0.58--0.72 pairwise accuracy and CLIPScore 0.55--0.62, while JPEGScore and, on BAGEL's own images, AestheticScore are near or below chance. The 95\% confidence intervals for QwenVQAScore's pairwise accuracy (percentile bootstrap over prompts) are $[0.743, 0.756]$, $[0.733, 0.760]$, and $[0.685, 0.823]$, and excluding the author annotator from (iii) gives 0.747 with the same ranking of rewards. QwenVQAScore is nonetheless an imperfect reward: in our human study, each annotator agrees with the other four (0.94 leave-one-out pairwise accuracy) more often than QwenVQAScore does, and on the detector-labeled set, thresholding QwenVQAScore at 0.5 misses more than half of the incorrect generations. We use QwenVQAScore as the reward for all subsequent experiments, with ImageReward results reported in Appendix \ref{app:imagereward_results} for comparison.

\begin{table}[t]
\centering
\small
\caption{Validation of reward functions against three independent references (Section~\ref{sec:rewards}). Within-prompt pairwise accuracy compares images of the same prompt, and AUROC pools images across prompts (0.5 = chance for both). The last row scores each human annotator against the other four.}
\label{tab:reward_validation}
\begin{adjustbox}{max width=\textwidth}
\begin{tabular}{lcccccc}
\toprule
& \multicolumn{3}{c}{Within-prompt pairwise accuracy} & \multicolumn{3}{c}{AUROC} \\
\cmidrule(lr){2-4} \cmidrule(lr){5-7}
Reward & \shortstack{GenEval\\detector} & \shortstack{GenAI-Bench\\humans} & \shortstack{Our human\\study} & \shortstack{GenEval\\detector} & \shortstack{GenAI-Bench\\humans} & \shortstack{Our human\\study} \\
\midrule
QwenVQAScore & \textbf{0.750} & \textbf{0.746} & \textbf{0.756} & \textbf{0.765} & \textbf{0.854} & \textbf{0.816} \\
ImageReward & 0.630 & 0.707 & 0.608 & 0.593 & 0.773 & 0.620 \\
PickScore & 0.577 & 0.718 & 0.651 & 0.611 & 0.770 & 0.655 \\
CLIPScore & 0.589 & 0.616 & 0.553 & 0.537 & 0.630 & 0.531 \\
AestheticScore & 0.507 & 0.630 & 0.517 & 0.497 & 0.626 & 0.514 \\
JPEGScore & 0.474 & 0.354 & 0.433 & 0.494 & 0.395 & 0.453 \\
\midrule
Human annotators (leave-one-out) & -- & -- & 0.944 & -- & -- & 0.978 \\
\bottomrule
\end{tabular}
\end{adjustbox}
\end{table}

% Per-prompt reward distributions and scores
\begin{figure}[]
    \centering
    % \includegraphics[width=0.8\textwidth]{figs/back_stuffie_reward.png}
    
    % \vspace{-0.25em}  % Reduce space between images
    \includegraphics[width=1.0\textwidth]{figs/text_cake_reward.pdf}
    \vspace{-0.5cm}
    \caption{Reward distributions for individual prompts show that QwenVQAScore effectively scores generations and produces intra-prompt rewards that distinguish good from bad samples; more prompt-level reward distributions are shown in Appendix \ref{app:reward_dists}. Table~\ref{tab:reward_validation} compares all rewards quantitatively against independent judgments of good and bad samples, including human annotations.}
    \label{fig:prompt_reward_dists}
    \vspace{-0.3cm}
\end{figure}

%% file: sec/modality_experiments.tex
% Alt titles: 
% Does Joint Text-Image Generation Improve Text-to-Image Performance?
%  The Impact of Text Generation in Multimodal Post-Training
% Evaluating the Contribution of Joint Text-Image Generation to T2I Performance
\section{The Impact of Joint Text-Image Generation on T2I Performance}\label{sec:modality_experiments}

In this set of experiments, we investigate two related questions: \textit{for a unified model that can generate in either mode, how much does joint text-image generation contribute to T2I performance, and which modality's loss should carry the reward signal?} 
Intuitively, using a unified model with strong language capabilities to generate text that guides image generation should improve T2I performance. However, multimodal models generally underperform image-only generation models, even on benchmarks requiring implicit knowledge and reasoning~\cite{wiseworldknowledgeinformedsemantic, qwenimagetechnicalreport, oneigbenchomnidimensionalnuancedevaluation}.

We address these questions through systematic modality ablations, comparing multimodal post-training against image-only post-training as well as reward-weighting each modality.
Both settings use the same images and reward labels: in the image-only setting the reasoning traces are removed, so there is no text loss, and the model generates without reasoning. 
We additionally compare modality importance in RWR by selectively reward-weighting different loss components: the text cross-entropy loss only (CE-weighted RWR), the image flow-matching loss only (MSE-weighted RWR), or both losses jointly (Multimodal RWR). These configurations differ only in where the reward weight is applied: in all three, the model is trained on the full reasoning-trace-and-image sequences with both losses active, all parameters are updated, and evaluation uses reasoning. In particular, MSE-weighted RWR is not an image-only method; the unweighted text loss on the reasoning traces is still optimized.
We train on our MMGW dataset discussed in Section \ref{sec:multimodal_gen_weaknesses_dataset}, with training hyperparameters in Appendix \ref{app:training_hyperparams}. 
All configurations train for the same number of steps with the same number of tokens per step. 

\subsection{Quantitative Results}\label{sec:modality_experiments_results}
\input{tables/geneval_modality}

\input{tables/dpg_modality}

We measure T2I performance across 4 benchmarks: GenEval~\cite{geneval}, an object-based prompt alignment benchmark; DPG-Bench~\cite{dpgbench}, which evaluates prompt alignment on dense prompts; WISE~\cite{wiseworldknowledgeinformedsemantic}, a benchmark requiring implicit understanding and world knowledge for successful generations; and OneIG-Bench~\cite{oneigbenchomnidimensionalnuancedevaluation}, a very difficult benchmark with fine-grained evaluation spanning prompt-image alignment, text rendering, reasoning-generated content, stylization, and diversity. Evaluation settings for all models are given in Appendix~\ref{app:eval_protocol}, along with how often the learned modality switch occurs, measured on our surviving OCR-reward checkpoint (Appendix~\ref{app:switching}) rather than the checkpoints reported in our tables. 

As seen in Table \ref{tab:geneval_modality}, Multimodal RWR achieves the best overall performance on GenEval. To gauge evaluation noise, we evaluated base BAGEL (image-only generation) on GenEval with 11 different sampling seeds, which gives overall scores from 0.769 to 0.790 (standard deviation 0.006). Multimodal SFT and the other RWR configurations are therefore comparable to BAGEL Multimodal (0.79--0.80 vs.\ 0.78). Among the loss-weighting configurations, reward-weighting both modalities is most effective.

On DPG-Bench (Table \ref{tab:dpgbench_modality}), Multimodal RWR scores 84.37 compared to 84.01 for the BAGEL Multimodal baseline. This difference of 0.36 points is comparable to the evaluation noise: re-scoring the same set of generated images with the benchmark's VQA evaluator changes the overall score by up to 0.3 points. We therefore consider Multimodal RWR comparable to the baseline on DPG-Bench rather than improved. The other multimodal post-trained configurations score slightly below the baseline (83.57--83.84), and image-only generation is strongest on this benchmark (85.08 for BAGEL Image-Only, and 86.98 and 86.75 for the image-only post-trained models; Section~\ref{sec:image_only}).

On WISE (Table \ref{tab:wise_modality}; GPT-4o judge), Multimodal RWR scores 0.72 compared with 0.70 for the BAGEL Multimodal baseline. This difference is similar to the spread among the other post-trained configurations (CE-weighted RWR, MSE-weighted RWR, and Multimodal SFT score 0.68--0.70), so we consider the two comparable. Among the post-trained configurations, reward-weighting both losses best preserves the base model's WISE performance. The largest effect on this knowledge-intensive benchmark is the generation mode: BAGEL's multimodal generation with reasoning substantially outperforms its image-only generation (0.70 vs.\ 0.52).

On OneIG-Bench (Table \ref{tab:oneigbench_modality}), out of all multimodal methods Multimodal RWR achieves the best Text Rendering score and competitive performance in Reasoning and Style evaluations. It raises the Text score of the BAGEL Multimodal Baseline from 0.020 to 0.189---an approximately ninefold improvement. Its Reasoning score (0.220) is the same as that of the other multimodal post-trained configurations, including Multimodal SFT (0.219--0.220).
Alignment and Style scores remain competitive with BAGEL Multimodal, though Diversity decreases, as expected since RWR post-trains exclusively on synthetic data and up-weights successful samples. Note that OneIG-Bench's Diversity definition only accounts for embedding space variance of intra-prompt images, with no consideration of prompt alignment or image quality. 
Relative to Multimodal SFT, all RWR variants improve text rendering---CE-weighted RWR 0.057, MSE-weighted RWR 0.077, and Multimodal RWR 0.189 compared to Multimodal SFT's 0.010---without sacrificing Alignment or Reasoning performance.
These results show that Multimodal RWR is the most effective configuration among the multimodal runs, improving text rendering while preserving broader capability. 
Overall on OneIG-Bench, image-only methods outperform multimodal, with stark decreases in performance when switching to multimodal generation even when using the same base model, which we discuss in Sections~\ref{sec:image_only} and~\ref{sec:multimodal_failures}.
\input{tables/wise_modality}
\input{tables/wise_common_judge}
\input{tables/oneig_modality}

\subsection{Is Reasoning Needed? Matched Image-Only Post-Training}\label{sec:image_only}
To test whether the improvements require intermediate reasoning, we train image-only counterparts of Multimodal SFT and Multimodal RWR on exactly the same MMGW images and reward labels, with the reasoning traces removed, and evaluate them without reasoning using the same generation settings as our other post-trained models (Image-Only SFT and Image-Only RWR in Tables~\ref{tab:geneval_modality}, \ref{tab:dpgbench_modality}, \ref{tab:oneigbench_modality}, and~\ref{tab:wise_common_judge}). These models never see reasoning in training and generate directly from the prompt, making them post-trained counterparts of BAGEL's image-only mode. Comparing them with Multimodal SFT and RWR isolates the contribution of intermediate reasoning, since images, reward labels, steps, and tokens per step are identical.

Comparing Image-Only RWR with Image-Only SFT shows that reward weighting also helps text rendering without reasoning: Image-Only RWR scores 0.475 on OneIG-Bench Text vs.\ 0.444 for Image-Only SFT, while the differences on GenEval (0.80 vs.\ 0.78), DPG-Bench, and the remaining OneIG-Bench metrics are small. Comparing the image-only and multimodal runs shows that whether reasoning helps depends on the benchmark. Image-only post-training outperforms multimodal post-training on dense-prompt alignment (DPG-Bench 86.75 for Image-Only RWR vs.\ 84.37 for Multimodal RWR) and on text rendering (OneIG-Bench Text 0.475 vs.\ 0.189), consistent with the analysis in Section~\ref{sec:multimodal_failures}. In contrast, reasoning-mode generation is substantially better on the knowledge-intensive WISE benchmark: under the common GPT-6.1 Sol judge, BAGEL Multimodal and our surviving OCR-reward checkpoint (Section~\ref{sec:rwr}; not the Multimodal RWR checkpoint in the other tables) both score 0.51, compared with 0.37--0.39 for image-only generation (Table~\ref{tab:wise_common_judge}). On GenEval, Multimodal RWR scores 0.83 and Image-Only RWR 0.80, and on OneIG-Bench Reasoning the two are comparable (0.220 vs.\ 0.210).

These results indicate that reward-weighted post-training is beneficial in both generation modes, and that intermediate reasoning helps most on prompts requiring world knowledge while hurting dense-prompt alignment and text rendering. Post-training in either mode also reduces OneIG-Bench Diversity, and Multimodal RWR's Style score is below that of BAGEL Multimodal (0.370 vs.\ 0.390).

%% file: tables/geneval_modality.tex
\begin{table*}[t]
% \vspace{-1.5em}
\centering
\small 
\caption{
    \textbf{Results on the GenEval Benchmark}; The best score is in \colorbox{best_result_color}{green} and the second-best score in \colorbox{second_best_result_color}{orange}. \multimodalbox{Blue border} indicates the best Multimodal score and \imageonlybox{purple border} indicates the best Image-Only score. $\dagger$ denotes base model baseline and $*$ denotes SFT baseline, trained on the same data as RWR but without reward-weighting.
}
% \vspace{-0.4cm}
\label{tab:geneval_modality}
\setlength{\tabcolsep}{3pt}  % Reduce cell padding
\begin{adjustbox}{max width=1.0\textwidth}
\begin{tabular}
{l@{\hspace{0.5cm}}l@{\hspace{0.5cm}}ccccccc}
\toprule
{\bf \shortstack{Modality\\Ablation}} & {\bf Model} & {\bf \shortstack{Single\\Object}}  & {\bf \shortstack{Two\\Object}} & {\bf Counting} & {\bf Colors} & {\bf Position} & {\bf \shortstack{Color\\Attribute}} & {\bf Overall $\uparrow$}
\\
\midrule

\multirow{5}{*}{Image-Only} 
& BAGEL Image-Only $^\dagger$~\cite{bagel} & 0.99 & 0.93 & 0.82 & 0.86 & 0.52 & 0.62 & 0.79 \\
& Image-Only SFT$^*$ & 1.00 & 0.95 & 0.76 & 0.88 & 0.46 & 0.65 & 0.78 \\
& Image-Only RWR & 1.00 & 0.93 & 0.81 & 0.88 & 0.51 & 0.68 & \imageonlyboxsecondbest{0.80} \\
& SD3-Medium~\cite{sd3_esser2024scalingrectifiedflowtransformers} & 0.99 & 0.94 & 0.72 & 0.89 & 0.33 & 0.60 & 0.74 \\
& FLUX.1[dev]~\cite{flux1kontextflowmatching} & 0.98 & 0.81 & 0.74 &  0.79 & 0.22 & 0.45 & 0.66 \\
\midrule

\multirow{7}{*}{Multimodal} 
& Show-o~\cite{show-o} & 0.98 & 0.80 & 0.66 & 0.84 & 0.31 & 0.50 & 0.68 \\
& Janus-Pro-7B~\cite{januspro} & 0.99 & 0.89 & 0.59 & 0.90 & 0.79 & 0.66 & \colorbox{second_best_result_color}{0.80} \\
& BAGEL Multimodal $^\dagger$~\cite{bagel} & 0.99 & 0.94 & 0.79 & 0.86 & 0.50 & 0.62 & 0.78 \\
& Multimodal SFT$^*$ & 0.99 & 0.93 & 0.79 & 0.86 & 0.54 & 0.64 & 0.79 \\
& CE-weighted RWR & 1.00 & 0.93 & 0.79 & 0.88 & 0.52 & 0.60 & 0.79 \\
& MSE-weighted RWR & 0.99 & 0.95 & 0.81 & 0.91 & 0.52 & 0.63 & \colorbox{second_best_result_color}{0.80} \\
& Multimodal RWR & 0.99 & 0.97 & 0.82 & 0.88 & 0.61 & 0.72 & \multimodalboxbest{0.83} \\

\bottomrule
\vspace{-0.5cm}
\end{tabular}
\end{adjustbox}
\end{table*}

%% file: tables/dpg_modality.tex
\begin{table*}[t]
% \vspace{-1.5em}
\centering
\small 
\caption{
    \textbf{Results on the DPG-Bench Benchmark}; Highlighting and markers as in Table~\ref{tab:geneval_modality}.
    % The best score is in \colorbox{best_result_color}{green} and the second-best score in \colorbox{second_best_result_color}{orange}. \multimodalbox{Blue border} indicates the best Multimodal score and \imageonlybox{purple border} indicates the best Image-Only score. $\dagger$ denotes base model baseline and $*$ denotes SFT baseline, trained on the same data as RWR but without reward-weighting.
}
% \vspace{-0.2cm}
\label{tab:dpgbench_modality}
\begin{adjustbox}{max width=1.0\textwidth}
\begin{tabular}
{ll@{\hspace{0.5cm}}cccccc}
\toprule
{\bf \shortstack{Modality\\Ablation} } & {\bf Model} & {\bf Global}  & {\bf Entity} & {\bf Attribute} & {\bf Relation} & {\bf Other} & {\bf Overall $\uparrow$}
\\
\midrule

\multirow{5}{*}{Image-Only} 
& BAGEL Image-Only$^{\dagger}$~\cite{bagel} & 90.73 & 90.05 & 89.72 & 92.23 & 90.80 & 85.08 \\
& Image-Only SFT$^{\ast}$ & 88.53 & 93.92 & 89.66 & 91.32 & 91.46 & \imageonlyboxbest{86.98} \\
& Image-Only RWR & 90.07 & 92.62 & 90.03 & 92.63 & 91.84 & \colorbox{second_best_result_color}{86.75} \\
& FLUX.1[dev]~\cite{flux1kontextflowmatching} & 74.35 & 90.00 & 88.96 & 90.87 & 88.33 & 83.84 \\
& SD3-Medium~\cite{sd3_esser2024scalingrectifiedflowtransformers} & 87.90 & 91.01 &  88.83 & 80.70 & 88.68 & 84.08 \\
\cmidrule{1-8}

\multirow{7}{*}{Multimodal} 
& BLIP3-o~\cite{blip3o} & - & - & - & - & - & 81.60 \\
& Janus-Pro-7B~\cite{januspro}& 86.90 & 88.90 & 89.40 & 89.32 & 89.48 & 84.19 \\
& BAGEL Multimodal$^{\dagger}$~\cite{bagel} & 88.75 & 90.84 & 89.01 & 89.44 & 89.70 & 84.01 \\
& Multimodal SFT$^{\ast}$ & 88.73 & 88.92 & 88.89 & 90.04 & 87.02 & 83.57 \\
& CE-weighted RWR & 89.94 & 89.10 & 89.24 & 89.41 & 90.58 & 83.84 \\
& MSE-weighted RWR & 90.39 & 90.87 & 88.74 & 87.67 & 86.97 & 83.60 \\
& Multimodal RWR & 89.20 & 90.63 & 89.73 & 90.03 & 86.81 & \multimodalbox{84.37} \\

\bottomrule
\vspace{-0.7cm}
\end{tabular}
\end{adjustbox}
\end{table*}

%% file: tables/wise_modality.tex
\begin{table}[t]
% \vspace{-1.5em}
\centering
\small
\caption{
    \textbf{Results on the WISE Benchmark \cite{wiseworldknowledgeinformedsemantic} (GPT-4o judge)}; All scores use the GPT-4o judge of the original WISE protocol; rows for BAGEL and prior models are taken from their papers. 
    The best score is in \colorbox{best_result_color}{green} and the second-best score in \colorbox{second_best_result_color}{orange}. \multimodalbox{Blue border} indicates the best Multimodal score and \imageonlybox{purple border} indicates the best Image-Only score. $\dagger$ denotes base model baseline and $*$ denotes SFT baseline, trained on the same data as RWR but without reward-weighting.
 $^{\S}$Image-Only SFT and RWR were evaluated after the GPT-4o judge became unavailable; their scores under the GPT-6.1 Sol judge are in Table~\ref{tab:wise_common_judge} and are not comparable to the GPT-4o scores here.
}
\label{tab:wise_modality}
\begin{adjustbox}{max width=1.0\textwidth}
\begin{tabular}
{l@{\hspace{0.5cm}}l@{\hspace{0.5cm}}ccccccc}
\toprule
\multirow{2}{*}{\bf \shortstack{Modality\\Ablation}} & \multirow{2}{*}{\bf Model}  & \multirow{2}{*}{\bf Cultural}  & \multicolumn{2}{c}{\bf Spatio-Temporal} & \multicolumn{3}{c}{\bf Natural Science } & {\multirow{2}{*}{\bf Overall $\uparrow$}}
\\
\cmidrule(lr){4-5}\cmidrule(lr){6-8}
&&&
{\bf Time} &
{\bf Space} &
{\bf Biology  } &
{\bf Physics} &
{\bf Chemistry} &
\\
\cmidrule{1-9}

% Image-Only models
\multirow{6}{*}{Image-Only} 
& BAGEL Image-Only$^\dagger$~\cite{bagel} & 0.44 & 0.55 & 0.68 & 0.44 & 0.60 & 0.39 & \imageonlybox{0.52} \\
& Image-Only SFT$^*$ & -- & -- & -- & -- & -- & -- & --$^{\S}$ \\
& Image-Only RWR & -- & -- & -- & -- & -- & -- & --$^{\S}$ \\
& FLUX.1[dev]~\cite{flux1kontextflowmatching} & 0.48 & 0.58 & 0.62 & 0.42 & 0.51 & 0.35 & 0.50 \\
& SD3.5 Large~\cite{sd3_esser2024scalingrectifiedflowtransformers} & 0.44 & 0.50 & 0.58 & 0.44 & 0.52 & 0.31 & 0.46 \\
& SD XL~\cite{sdxlimprovinglatentdiffusion} & 0.43 & 0.48 & 0.47 & 0.44 & 0.45 & 0.27 & 0.43 \\
\cmidrule{1-9}

% Multimodal models
\multirow{9}{*}{Multimodal} 
& BLIP3-o~\cite{blip3o} & - & - & - & - & - & - & 0.62 \\
& T2I-R1~\cite{t2ir1reinforcingimagegeneration} & 0.56 & 0.55 & 0.63 & 0.54 & 0.55 & 0.30 & 0.54 \\
& Show-o~\cite{show-o} & 0.28 & 0.40 & 0.48 & 0.30 & 0.46 & 0.30 & 0.35 \\
& Janus-Pro-7B~\cite{januspro}& 0.30 & 0.37 & 0.49 & 0.36 & 0.42 & 0.26 & 0.35 \\
& BAGEL Multimodal$^\dagger$~\cite{bagel} & 0.76 & 0.69 & 0.75 & 0.65 & 0.75 & 0.58 & \colorbox{second_best_result_color}{0.70} \\
& Multimodal SFT$^*$ & 0.70 & 0.64 & 0.72 & 0.60 & 0.77 & 0.65 & 0.68 \\
& CE-weighted RWR & 0.73 & 0.65 & 0.75 & 0.60 & 0.80 & 0.63 & 0.69 \\
& MSE-weighted RWR & 0.73 & 0.67 & 0.74 & 0.64 & 0.80 & 0.65 & \colorbox{second_best_result_color}{0.70} \\
& Multimodal RWR & 0.76 & 0.69 & 0.77 & 0.64 & 0.77 & 0.66 & \multimodalboxbest{0.72} \\

\bottomrule
\end{tabular}
\end{adjustbox}
\end{table}

%% file: tables/wise_common_judge.tex
% Rebuttal: WISE under a common judge (GPT-6.1 Sol), so image-only post-trained models can be compared with reasoning-mode generation.
\begin{table}[t]
\centering
\small
\caption{\textbf{WISE under a common judge (GPT-6.1 Sol).} Because the original GPT-4o judge is no longer available, all rows are scored with GPT-6.1 Sol using the original WISE prompt and scoring. Scores are lower than, and not comparable to, those in Table~\ref{tab:wise_modality}. Image-only rows use the generation settings of our post-trained models (Appendix~\ref{app:eval_protocol}), and BAGEL Multimodal uses BAGEL's thinking pipeline. $^\ddagger$Not the Multimodal RWR checkpoint reported in Table~\ref{tab:wise_modality} and the other main tables: this is our only surviving post-trained multimodal checkpoint, a Multimodal RWR variant whose reward additionally includes an OCR score on long-text prompts, also used in Appendices~\ref{app:eval_protocol} and~\ref{app:switching}. Its learned modality switch occurs on 33\% of WISE prompts, and the remaining images come from the manual modality switch.}
\label{tab:wise_common_judge}
\resizebox{0.85\textwidth}{!}{
\begin{tabular}{l@{\hspace{0.4cm}}l@{\hspace{0.4cm}}ccccccc}
\toprule
{\bf Mode} & {\bf Model} & {\bf Cultural} & {\bf Time} & {\bf Space} & {\bf Biology} & {\bf Physics} & {\bf Chemistry} & {\bf Overall $\uparrow$} \\
\midrule
\multirow{3}{*}{Image-Only}
& BAGEL Image-Only$^\dagger$ & 0.33 & 0.39 & 0.47 & 0.30 & 0.41 & 0.32 & 0.37 \\
& Image-Only SFT$^*$ & 0.34 & 0.40 & 0.47 & 0.32 & 0.43 & 0.35 & 0.38 \\
& Image-Only RWR & 0.35 & 0.40 & 0.52 & 0.29 & 0.45 & 0.36 & 0.39 \\
\midrule
\multirow{2}{*}{Multimodal}
& BAGEL Multimodal$^\dagger$ & 0.58 & 0.50 & 0.60 & 0.45 & 0.51 & 0.45 & 0.51 \\
& Multimodal RWR (+OCR reward)$^\ddagger$ & 0.57 & 0.50 & 0.56 & 0.43 & 0.51 & 0.46 & 0.51 \\
\bottomrule
\end{tabular}
}
\end{table}

%% file: tables/oneig_modality.tex
\begin{table}[t]
% \vspace{-1.5em}
\centering
\small
\caption{
    \textbf{Results on the OneIG-Bench Benchmark}; 
    % The best score is in \colorbox{best_result_color}{green} and the second-best score in \colorbox{second_best_result_color}{orange}. $\dagger$ denotes base model baseline. $*$ denotes SFT baseline, trained on the same data as RWR but without reward-weighting.
}
\label{tab:oneigbench_modality}
\begin{adjustbox}{max width=1.0\textwidth}
\begin{tabular}
{l@{\hspace{0.5cm}}l@{\hspace{0.5cm}}ccccc}
\toprule
{\bf \shortstack{Modality\\Ablation}} & {\bf Model} & {\bf Alignment $\uparrow$}  & {\bf Text $\uparrow$} & {\bf Reasoning $\uparrow$} & {\bf Style $\uparrow$} & {\bf Diversity $\uparrow$}
\\
\midrule

\multirow{6}{*}{Image-Only}
& BAGEL Image-Only$^\dagger$~\cite{bagel} & 0.769 & 0.244 & 0.173 & 0.367 & 0.251 \\
& Image-Only SFT$^*$ & \imageonlyboxbest{0.837} & 0.444 & 0.209 & 0.374 & 0.179 \\
& Image-Only RWR & \imageonlyboxbest{0.837} & 0.475 & 0.210 & \imageonlyboxsecondbest{0.377} & 0.184 \\
& FLUX.1[dev]~\cite{flux1kontextflowmatching} & 0.786 &  \colorbox{second_best_result_color}{0.523} & \colorbox{second_best_result_color}{0.253} & 0.368 & 0.238 \\
& SD XL~\cite{sdxlimprovinglatentdiffusion} & 0.688 & 0.029 & 0.237 & 0.332 & \imageonlyboxsecondbest{0.296} \\
% SD3.5 Large is #1 in Alignment, Text, Reasoning !!!
& SD3.5 Large~\cite{sd3_esser2024scalingrectifiedflowtransformers} & 
\colorbox{second_best_result_color}{0.809} & \imageonlyboxbest{0.629} & \imageonlyboxbest{0.294} & 0.353 & 0.225 \\
\cmidrule{1-7}

\multirow{8}{*}{Multimodal} 
& Show-o~\cite{show-o} & 0.702 & 0.002 & 0.213 & 0.361 & 0.241 \\
& BLIP3-o~\cite{blip3o} & 0.711 & 0.013 & \multimodalbox{0.223} & 0.361 & 0.229 \\
& Janus-Pro-7B~\cite{januspro}& 0.553 & 0.001 & 0.139 & 0.276 & \multimodalboxbest{0.365} \\
& BAGEL Multimodal$^\dagger$~\cite{bagel} & 0.793 & 0.020 & 0.206 & \multimodalboxbest{0.390} & 0.209 \\
& Multimodal SFT$^*$ & 0.800 & 0.010 & 0.219 & 0.365 & 0.143 \\
& CE-weighted RWR & 0.803 & 0.057 & 0.220 & 0.366 & 0.149 \\
& MSE-weighted RWR & \multimodalbox{0.804} & 0.077 & 0.219 & 0.364 & 0.144 \\
& Multimodal RWR & 0.802 & \multimodalbox{0.189} & 0.220 & 0.370 & 0.141 \\

\bottomrule
\end{tabular}
\end{adjustbox}
\end{table}

%% file: sec/multimodal_failures.tex
% When Multimodal Generation Helps (or Hurts) Text-to-Image Performance
\subsection{When Multimodal Generation Harms Text-to-Image Performance}\label{sec:multimodal_failures}

% Example Failure
\begin{figure}[h]
  \centering
  \includegraphics[width=0.9\textwidth]{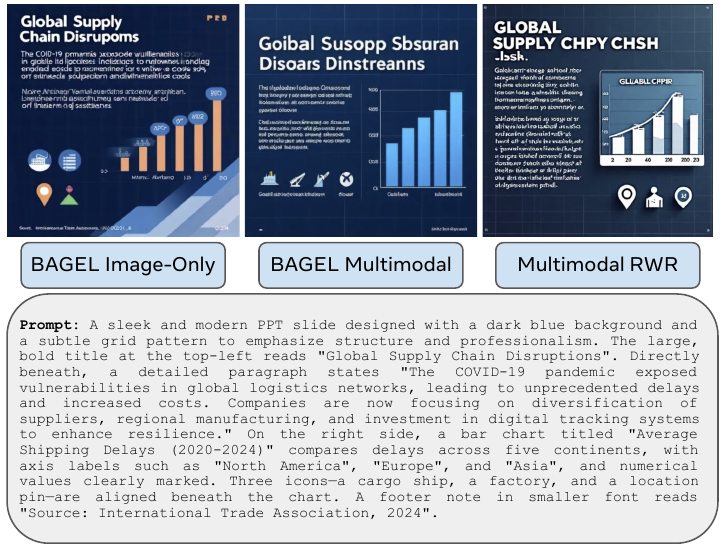}
  \vspace{-0.2cm}
  \caption{
  Sample generations for a OneIG-Bench Text category prompt. BAGEL Image-Only produces the clearest text, consistent with the stronger text accuracy of image-only generation methods seen in Table \ref{tab:oneigbench_modality}. Our Multimodal RWR post-training helps remedy some of the text rendering failures of BAGEL Multimodal. 
  }
  \label{fig:text_failure_gen}
  \vspace{-0.5cm}
\end{figure}

% In this section, we discuss when multimodal generation notably degrades T2I performance, compared to image-only generation.
The Text category of OneIG-Bench (Table \ref{tab:oneigbench_modality}) exposes that the performance of Image-Only models is generally much higher than Multimodal models; specifically for BAGEL, Multimodal generation results in a roughly 12x performance degradation compared to Image-Only generation.
Multimodal RWR greatly remedies this performance drop, achieving a Text score of 0.189 compared to BAGEL Multimodal's 0.020, but cannot fully close the gap to BAGEL Image-Only (0.244).
These results indicate that text rendering is a specific area in which text-image generation worsens performance compared to image-only generation. 

The Text category prompts in OneIG-Bench are highly challenging, requiring generating complex images that accurately render multiple text phrases~\cite{oneigbenchomnidimensionalnuancedevaluation}; Figure~\ref{fig:text_failure_gen} shows an example prompt and generations.
% with generations from BAGEL Image-Only, BAGEL Multimodal, and BAGEL after Multimodal RWR post-training.
One possible factor behind image-only generation outperforming text-image generation on text rendering tasks is that conditioning on a reasoning trace may overload the model with excessive textual input, as reasoning can guide text placement and scene layout but does not directly improve text rendering. Reasoning traces of our surviving OCR-reward checkpoint (Section~\ref{sec:rwr}; not the Multimodal RWR checkpoint in Table~\ref{tab:oneigbench_modality}) point to a related factor: for OneIG-Bench text prompts on which the model switches to image generation itself, the reasoning reproduces all of the prompt's quoted text strings in only 36\% of cases (53\% of strings on average), often paraphrasing them into a longer scene description. Because image generation in this setting is conditioned on the reasoning, omitted or altered strings cannot be rendered correctly.
This can be seen in Figure \ref{fig:text_failure_gen}, as BAGEL Image-Only's generation contains the clearest text but a more visually cluttered composition compared to the multimodally generated images.
Overall these results highlight that although multimodal reward weighting can address some gaps, image-only models still achieve the best performance on certain tasks.
This suggests there is still substantial room for improving how reasoning conditions image generation. 
% \yushi{Another reason might be that QwenVQAScore is not a good reward that can capture text rendering errors? So it might be possible to fix this if we have better reward models.}\MG{But Bagel + thinking without reward function is really bad, so the reason is not Qwen.}
% Overall on OneIG-Bench, Image-Only models---particularly SD3.5 Large---attain the highest scores in Alignment, Text Rendering, and Reasoning capabilities.

% DPG-Bench results in Table \ref{tab:dpgbench_modality} show that BAGEL's Image-Only generation outperforms all multimodal generation methods. 
% Multimodal RWR is the best multimodal method, demonstrating a clear improvement over the corresponding BAGEL Multimodal baseline and narrowing the performance gap to the Image-Only baseline while outperforming both SFT settings and prior multimodal models across several subcategories. This suggests that while Image-Only generation remains advantaged on DPG-Bench, reward-weighting both modalities can yield measurable gains.

%% file: sec/dataset_experiments.tex
\section{Post-Training Data Strategies}\label{sec:data_strategies_experiments}
\input{tables/all_benchmarks_datasets}

Our approach uses training prompts from semantic categories that elicit inconsistent generation quality from the base model, where only a subset of samples are successful generations. 
This contrasts with two common strategies: (1) using general internet-scale image--caption datasets for broad generalization, and (2) employing benchmark-specific prompts that mirror evaluation sets, as in recent online RL work~\cite{flowgrpotrainingflowmatching}, which optimizes benchmark scores but risks overfitting to the task and prompt design.

To compare these data strategies to our weakness-targeted approach, we run our synthetic data generation and RWR training on three different training prompt sets: (1) captions from a licensed general image-caption dataset (Shutterstock); (2) the GenEval-generated prompts from FlowGRPO~\cite{flowgrpotrainingflowmatching}, constructed using the benchmark's prompt templates, which serve as an in-distribution reference for GenEval because they share its templates and object vocabulary (73\% of GenEval prompts have a training prompt with exactly the same objects); (3) our MMGW prompt set described in Section \ref{sec:multimodal_gen_weaknesses_dataset}.
All data ablations use an identical budget of 357,300 training samples, consisting of 100 samples of text reasoning and images generated per prompt and reward-labeled with QwenVQAScore.
Although matched in size, the prompt sets differ substantially in the reward distributions they induce. Counting a generation as successful when its QwenVQAScore reward is at least 0.5, only 4.7\% of MMGW prompts succeed on every sample, compared to 20.4\% of GenEval-generated prompts; because reward weighting can only exploit differences among samples of the same prompt, such saturated prompts contribute little training signal.
Mean intra-prompt reward standard deviation is correspondingly higher for MMGW, 0.254 versus 0.172 for GenEval-generated prompts.

The prompt sets also differ in their semantic coverage. MMGW has the most text-rendering prompts (32\% of prompts, versus 0\% for GenEval-generated prompts and 4\% for Shutterstock captions), and 23\% of its prompts involve counting (4\% for the others). 
To measure semantic overlap with the evaluation benchmarks, we embed every training and benchmark prompt and compute each benchmark prompt's maximum cosine similarity to each training set (Appendix~\ref{app:prompt_overlap}). No training prompt duplicates a benchmark prompt. As intended by their construction, GenEval-generated prompts are the closest to GenEval. Among all benchmark categories, MMGW overlaps most with OneIG-Bench's text-rendering category, since it has the largest share of text-rendering prompts; on DPG-Bench, WISE, and the remaining OneIG-Bench categories, its overlap is similar to that of Shutterstock captions. Because the prompt sets differ in reward variance, category composition, and overlap at the same time, our comparison does not isolate the effect of targeting model weaknesses, and we draw conclusions only about the specific prompt sets tested.
All training runs use the best-performing post-training method from Section \ref{sec:modality_experiments}, Multimodal RWR, with the same training settings, and we evaluate performance on the same four benchmarks.

All results are presented in Table \ref{tab:all_results_datasets}, with fine-grained results for each benchmark in Appendix \ref{app:full_dataset_results}.
GenEval~\cite{geneval} results show that while the GenEval-generated dataset predictably improves performance over both the base model and the Shutterstock dataset, our MMGW dataset achieves the highest score (0.83 vs.\ 0.82), although the difference from the in-distribution GenEval-generated prompts is small. 
% Notably, post-training on the Shutterstock dataset yields no performance improvements on GenEval, indicating that for a saturated benchmark with relatively simple prompts, prompts from general image-caption datasets may not be effective without more sophisticated post-training. 
On DPG-Bench~\cite{dpgbench}, MMGW (84.37), Shutterstock (84.23), and the base model (84.01) differ by amounts comparable to the evaluation noise (Section~\ref{sec:modality_experiments_results}), while performance degrades when using GenEval-generated training prompts---an expected outcome given that DPG-Bench's prompts are approximately four times longer than GenEval's and contain thousands of unique nouns compared to GenEval's 80 object classes.

For T2I benchmarks requiring reasoning or world knowledge, our MMGW dataset continues to outperform the other data strategies. 
On WISE~\cite{wiseworldknowledgeinformedsemantic} (GPT-4o judge), MMGW enables the best overall T2I performance. 
Notably, post-training on the Shutterstock dataset causes WISE performance to degrade compared to the base model, indicating that general image-caption prompts may be too simple and out-of-distribution for image generation tasks requiring implicit knowledge integration.   
On OneIG-Bench~\cite{oneigbenchomnidimensionalnuancedevaluation}, training on MMGW results in the best performance compared to other data strategies on all evaluation axes except for Diversity. MMGW's targeted prompts notably improve performance on the very difficult Text category, recovering much of the 12x performance drop of BAGEL's multimodal generation compared to BAGEL's image-only generation, whereas all other data strategies cause Text category performance to degrade further. Since MMGW contains far more text-rendering prompts than the other prompt sets, this gain reflects coverage of the category as much as the targeting of a weakness.

%% file: tables/all_benchmarks_datasets.tex
\begin{table*}[t]
\centering
\small
\caption{
    \textbf{Results Across All Benchmarks for Data Strategy Ablations}; The best score for each benchmark is in \colorbox{best_result_color}{\textbf{bold green}} and the second-best score in \colorbox{second_best_result_color}{\textit{italic orange}}. WISE scores use the GPT-4o judge. 
}
% \vspace{-0.3cm}
\label{tab:all_results_datasets}
\resizebox{1.0\textwidth}{!}{
\begin{tabular}{l@{\hspace{0.3cm}}c@{\hspace{0.3cm}}c@{\hspace{0.3cm}}c@{\hspace{0.3cm}}ccccc}
\toprule
\multirow{2}{*}{\bf \shortstack{Training Prompt\\Set Ablation}} & 
\multirow{2}{*}{\bf \shortstack{GenEval\\Overall $\uparrow$}} & 
\multirow{2}{*}{\bf \shortstack{DPG-Bench\\Overall $\uparrow$}} & 
\multirow{2}{*}{\bf \shortstack{WISE (GPT-4o)\\Overall $\uparrow$}} & 
\multicolumn{5}{c}{\bf OneIG-Bench}
\\
\cmidrule(lr){5-9}
& & & & 
{\bf Alignment $\uparrow$} & 
{\bf Text $\uparrow$} & 
{\bf Reasoning $\uparrow$} & 
{\bf Style $\uparrow$} & 
{\bf Diversity $\uparrow$}
\\
\midrule

Shutterstock & 
0.79 & 
84.23 & 
0.66 & 
\colorbox{second_best_result_color}{\textit{0.801}} & 
0.008 & 
\colorbox{second_best_result_color}{\textit{0.219}} & 
0.369 & 
0.145 
\\

GenEval-Generated~\cite{flowgrpotrainingflowmatching} & 
\colorbox{second_best_result_color}{\textit{0.82}} & 
83.16 & 
\colorbox{second_best_result_color}{\textit{0.70}} & 
0.797 & 
0.006 & 
0.217 & 
0.358 & 
0.147 
\\

MMGW (Ours) & 
\colorbox{best_result_color}{\textbf{0.83}} & 
\colorbox{second_best_result_color}{\textit{84.37}} & 
\colorbox{best_result_color}{\textbf{0.72}} & 
\colorbox{best_result_color}{\textbf{0.802}} & 
\colorbox{second_best_result_color}{\textit{0.189}} & 
\colorbox{best_result_color}{\textbf{0.220}} & 
\colorbox{second_best_result_color}{\textit{0.370}} & 
0.141 
\\

\midrule

BAGEL Multimodal (Baseline)~\cite{bagel} & 
0.78 & 
84.01 & 
\colorbox{second_best_result_color}{\textit{0.70}} & 
0.793 & 
0.020 & 
0.206 & 
\colorbox{best_result_color}{\textbf{0.390}} & 
\colorbox{second_best_result_color}{\textit{0.209}} 
\\

BAGEL Image-Only (Baseline)~\cite{bagel} & 
0.79 & 
\colorbox{best_result_color}{\textbf{85.08}} & 
0.52 & 
0.769 & 
\colorbox{best_result_color}{\textbf{0.244}} & 
0.173 & 
0.367 & 
\colorbox{best_result_color}{\textbf{0.251}} 
\\

\bottomrule
\vspace{-0.8cm}
\end{tabular}
}
\end{table*}

%% file: sec/conclusion.tex
\section{Conclusion and Future Work}
\vspace{-0.1cm}
In this work, we investigated offline reward-weighted post-training for reasoning-conditioned image generation in BAGEL, where the model also learns to emit the image-start token after its reasoning.
Applying Multimodal Reward-Weighted Regression (RWR), which integrates offline reward signals into both text and image losses, improved object-centric prompt alignment and text rendering over the multimodal baseline, with DPG-Bench and WISE performance comparable to the baseline.
We further find that the choice of post-training prompts strongly influences outcomes: among the prompt sets we tested, our weakness-targeted set (MMGW) produces the largest improvements. However, it also differs from the alternatives in reward variance, category composition, and semantic overlap with the benchmarks, so these gains cannot be attributed to weakness targeting alone (see Limitations).
Combining Multimodal RWR with MMGW gives the best GenEval and OneIG-Bench text-rendering scores among the multimodal configurations we evaluate. Matched image-only post-training remains stronger on dense-prompt alignment and text rendering, whereas reasoning-mode generation is markedly better on knowledge-intensive prompts.
Finally, reward weighting improves over SFT on the same data in both generation modes, most consistently in text rendering, underscoring the importance of how the reward is applied during post-training.

\vspace{-0.5cm}
\paragraph{Limitations and Future Work}
 As shown in Section~\ref{sec:multimodal_failures}, image-only generation currently outperforms multimodal generation on text rendering tasks; while our approach narrows this gap, unified text--image generation is not yet fully competitive. Post-training also reduces output diversity in both generation modes and lowers OneIG-Bench Style for multimodal generation.
 Promising future directions include analyzing when joint text--image generation helps versus hurts T2I performance, and post-training models to adaptively select the better strategy.
 Our study is also confined to a single unified architecture, BAGEL, which pairs autoregressive text generation with flow-matching image synthesis; extending this analysis to fully autoregressive models such as Janus-Pro~\cite{januspro} or discrete masked-token models such as Show-o~\cite{show-o}, where the text and image objectives take a different form, would establish how broadly these conclusions hold.
 Finally, we use a single reward model and a single round of sampling, reward labeling, and training; whether gains compound over further rounds, or hold under a different reward, remains open.
 Our data-strategy comparison matches the number of training samples but not reward variance, category composition, or semantic overlap with the benchmarks; controlled comparisons (e.g., reward-variance-matched subsets, or MMGW without text-rendering prompts) would be needed to attribute the gains to weakness targeting specifically.

%% file: sec/X_suppl.tex
\section{MMGW Dataset Prompt Subset}\label{app:mmgw_subset}
We provide $\sim$50 prompts from each of the 5 categories of our Multimodal Generative Weaknesses (MMGW) dataset used in post-training. 
These prompts were manually verified to yield generation failures in the majority of samples when using leading multimodal generative models, including GPT-4o---and we use them as examples to LLM-generate our full MMGW prompt dataset. 
Our main paper results show that synthetic datasets generated using MMGW prompts yielded effective post-training gains. 

\begin{tcolorbox}[promptbox,title={Relative Positions}]
    An orange cat sitting in front of a coffee cup on a table,
    A cat on top of a bird,
    A gnome behind a hut,
    A tropical cocktail on the left of flip-flops on a beach,
    A dog lying down to the right of some socks,
    A red apple placed behind a glass of water,
    A teddy bear sitting to the left of a stack of books,
    A sponge on top of a bottle of dish soap,
    A bicycle behind a tree,
    A pair of sunglasses under a beach towel,
    A hair clip to the right of a hairbrush,
    A moose walking to the right of a squirrel,
    A bird standing at the base of a fountain,
    A hat under a pile of scarves,
    A husky resting behind a sled,
    A horse standing behind a tractor,
    A herd of sheep behind a barn,
    Deer behind a campsite,
    Squirrels behind a picnic table,
    A book inside a breifcase,
    A painting resting on the ground below a shelf,
    A fish swimming behind a rock in an aquarium,
    Wolves behind a winter cabin,
    A kite flying to the left of a seagull,
    Candy scattered outside of an empty jar,
    Paper towels to the right of dirty dishes,
    A bird standing on top of a birdhouse,
    A bar of soap on top of a sink,
    An elf hiding behind a rock,
    Eggs outside of an empty carton,
    A guitar leaning against the left side of a wall,
    A bunny hiding behind a bush,
    Milk cartons behind a cat,
    A clock on the wall under a shelf,
    A book on a shelf to the right of a globe,
    A marlin swimming behind a school of fish,
    A pair of gloves to the right of some boots,
    A bicycle parked to the left of a lamppost,
    Clown fish swimming behind a sea anemone,
    A dolphin jumping out of the water in front of a boat,
    A herding dog behind of a flock of sheep,
    A rose growing to the left of a fire hydrant,
    A picture frame resting on top of a box,
    Pigs behind a chicken coop,
    A herd of water buffalo behind a jeep,
    A turtle basking behind some ducks,
    A book on a table to the left of a cup of tea,
    A bow to the right of a pair of heels,
    A hairbrush hanging in front of a mirror,
    An ant stuck under an apple,
    A purse behind a vase of flowers
\end{tcolorbox}

\begin{tcolorbox}[promptbox,title={Object Orientations}]
    A motorcycle lying down,
    An upside down water bottle,
    A bell pepper lying on its side,
    A spider flipped over,
    A chair turned upside down,
    A jug of milk knocked over on its side,
    A spoon resting on its side,
    A bunch of carrots hanging upside down,
    A perfume bottle lying flat,
    A computer monitor knocked over, laying flat,
    A hat placed upside down,
    A calculator lying face down,
    A beetle stuck upside down,
    A bicycle lying on its side,
    A mug standing upside down,
    A plate balanced on its rim,
    A clock face down on a table,
    A chair leaning against a wall,
    A vase lying on its side,
    A camera placed upside down,
    A pill bottle standing upside down,
    A suitcase lying flat,
    A lamp knocked over on its side,
    A skateboard flipped over, lying upside down,
    A basket turned upside down,
    A picture frame lying flat,
    A cup standing upside down on its rim,
    An open book lying face down on its pages,
    A chair turned on its side,
    A computer mouse placed upside down,
    A plate lying face down,
    A teddy bear lying face down,
    A fish swimming upside down,
    A whale swimming up towards the ocean surface,
    A crab flipped upside down,
    A daisy lying face down,
    A ring rolling along its edge,
    An upside down tomato resting on its stem,
    A pinneapple lying on its side,
    A wine glass lying on its side,
    Roller skates lying flat on their side,
    A keyboard lying upside down,
    A jellyfish floating upside down,
    A television lying flat on its face on the ground,
    A coin standing up on its edge,
    An upside down toaster,
    A stool turned upside down,
    A turtle flipped over on its shell,
    A penguin lying flat on its belly,
    An upside down tissue box,
    A pair of boots knocked down lying flat
\end{tcolorbox}

\begin{tcolorbox}[promptbox,title={Text}]
    A book page showing a photo of a jaguar swimming with the caption "Jaguars are great swimmers.",
    An astronaut floating in space has a speech bubble that says "Earth is so small from here",
    An angry dog frowning and saying "You fed me less today!",
    A still from a movie scene with the subtitles "It was all in his head.",
    A minimalistic invitation card that reads "Join us for our destination wedding" with a small palm tree icon.,
    A minimalistic card that reads "Missing you, get well soon!" with a bandaged cat.,
    A movie poster with the title "Zombie Cats" showing menacing cartoon cats, with a subtitle that says "The purr-fect horror comedy!",
    A gaming poster titled "Galactic Quest" showing a space adventurer with the subtitle "Embark on an interstellar journey.",
    A children's adventure book titled "Jungle Quest" with explorers and the subtitle "Thrills in the wild.",
    A large hospital with the name "Houston Medical Complex" is in a busy downtown area,
    A cozy tavern in a medieval village with a sign that reads "The Drunken Dragon",
    A highway sign showing 60 MPH,
    A digital alarm clock showing 6am,
    A comic book character saying "hello",
    A magazine cover with the title: Home Interiors,
    A book cover with the title: "Animals",
    A billboard showing the words: "Sale",
    A street sign showing "Missiont St",
    A restaurant sign with the title "Mary's",
    A website banner reading "Breaking News",
    A poster with the words "Live Concert Tonight",
    A chalkboard with the phrase "Today's Special",
    A coffee cup with the words "Good Morning",
    A banner showing "Happy Holidays",
    A postcard with the message "Wish You Were Here",
    A namecard with the name "Sally Smith",
    A price tag with the number "30\$",
    A book spine with the title "Mystery Novel",
    A cereal box with the label "Whole Grain",
    A wine label with the year "2015",
    A movie ticket with the time "7:00 PM",
    A greeting card with the words "Happy Birthday",
    A jar with a label reading "Milk",
    A label with the text "Flour",
    A sign with the words "No Parking",
    A magazine cover titled "World Travel",
    A banner with the title "Charity Run",
    A notebook with the word "Journal",
    A label on a jar reading "Peanut Butter",
    A road sign with the distance "10 Miles",
    A book cover with the author "Jane Austen",
    A menu board with the drink "Latte",
    A poster with the date "July 4th",
    A t-shirt with the phrase "Friyay",
    A classroom whiteboard with "Algebra" written on it,
    A poster with the text: "Move Fast",
    A postcard with the location "Paris",
    A name plate that says "CEO",
    A baseball cap with the letters "ATL",
    A book spine showing the author "Mark Twain",
    A cereal box with the word "Organic",
    A wine label with the region "Napa Valley",
    A sign advertising a movie titled "The Titanic",
    A greeting card with the phrase "Congratulations!",
    A sign in a store that says "Eggs",
    A sign with the words "Exit Only",
    A magazine cover titled "Health and Fitness",
    A sidewalk sign titled: "Art Exhibition",
    A soda can with the label "Zest" on it,
    A neon sign reading "brunch",
    A label on a jar reading "Honey",
    A book cover with the title "Adventure Stories"
\end{tcolorbox}

\begin{tcolorbox}[promptbox,title={Cardinality}]
    A beach with 4 palm trees,
    5 birds sitting on a power line,
    5 trees in an field,
    A tray of 6 cookies,
    A box of 6 markers,
    5 ducks swimming in a pond,
    A cup of 3 pens on a desk,
    4 geese swimming in a lake,
    A box of 6 crayons,
    4 pencils in a cup,
    3 crabs on a coral reef,
    6 flowers in a vase,
    A fruit basket holding 3 peaches,
    A crate holding 3 watermelons,
    A poster with 5 hearts on it,
    6 fish in an aquarium,
    An office with 3 plants,
    3 slices of pie on a plate,
    10 birds flying in formation,
    A school of 8 fish in the ocean,
    A bedroom with 2 houseplants,
    4 glasses of champaigne on a platter,
    An egg carton holding 6 eggs,
    A box of 4 soda cans,
    A herd of 5 bison,
    A highway with 4 sportscars on it,
    4 bowls in a kitchen cupboard,
    6 candles on a cake,
    4 acorns on a branch,
    A poster with 8 types of food on it,
    6 glasses in a kitchen cupboard,
    4 towels in a bathroom,
    5 keys on a keyring,
    A building with 6 windows,
    A classroom with 4 desks,
    A herd of 6 sheep,
    4 sandwhiches in a deli case,
    A flag with 10 stars,
    A poster with 5 plants on it,
    A fruit basket holding 3 bananas,
    A shelf with 5 candles,
    A poster showing 5 planes,
    A village with 5 huts,
    3 houseplants in a living room,
    A stack of 3 pancakes,
    A trail of 5 ants,
    5 stools at a bar,
    A crate holding 6 tomatoes,
    A label with 4 stars,
    A bakery display case containing 4 cupcakes,
    6 popsicles in a freezer
\end{tcolorbox}

\begin{tcolorbox}[promptbox,title={Structural Characteristics}]
    Fresh fruits all cut in half,
    A piece of paper folded in half,
    A bent knife,
    A car with a dented door,
    A deflated tire,
    A butterfly with its wings closed,
    A mug with a missing handle,
    A piece of fabric torn in half,
    A cracked phone screen,
    A chair with a missing leg,
    A warped tabletop,
    A television broken in half,
    A faded photograph of a dog,
    A scratched CD,
    A crushed, flattened water bottle,
    A flattened cardboard box,
    A flattened soda can,
    A dented sheild,
    A hummingbird with its wings closed,
    A watermelon sliced in half,
    A carrot cut in half,
    A wooden bridge with splinters,
    A cake cut in half,
    A book torn in half,
    A wooden table with big scratches,
    A scratched couch,
    A laptop broken in half,
    A bent needle,
    A bent paperclip,
    A cucumber cut in half,
    A tree branch broken in half,
    A table with a missing leg,
    Drywall with a hole in it,
    Socks turned inside out,
    A t-shirt torn in half,
    Celery sliced in half,
    A pencil snapped in half,
    A bent fork,
    Pants turned inside out,
    An action figure broken in half,
    A cupcake with no frosting',
    A scratched painting on a wall,
    A boat broken in half,
    A frayed wire,
    A dented trash can,
    A gummy bear cut in half,
    A plastic bag torn in half,
    A bent nail,
    A cracked tile,
    A chair with the legs broken off,
    A teddy bear missing an arm
\end{tcolorbox}
    
\clearpage

%% Full results for Dataset Experiments
\section{Detailed Results for Data Strategies Experiments in Section \ref{sec:data_strategies_experiments}}
\label{app:full_dataset_results}

\input{tables/geneval_dataset}
\input{tables/dpg_dataset}
\input{tables/wise_datasets}
\input{tables/oneig_datasets}

\clearpage

%% Evaluation protocol (rebuttal)

\section{Evaluation Protocol}
\label{app:eval_protocol}
\paragraph{Post-trained multimodal models.} All post-trained multimodal models (Multimodal, CE-weighted, and MSE-weighted RWR, Multimodal SFT, rejection sampling, and the data-strategy and ImageReward runs) are evaluated with the non-EMA weights of the final checkpoint using the learned image-start token. The benchmark prompt is given without a system prompt, and the model decodes greedily for up to 1,000 tokens. If it emits \vstarttok{}, the image is generated conditioned on the generated reasoning (Section~\ref{sec:rwr}) at 512$\times$512 with a 50-step flow-matching schedule, timestep shift 1.0, and text classifier-free guidance of scale 10 over the full trajectory, without image guidance. If no image-start token is emitted, GenEval and WISE first retry once with ``Generate'' prepended to the prompt. If the token is still not emitted, and on the other benchmarks, the evaluation falls back to a manual modality switch using BAGEL's standard thinking pipeline. In this pipeline, reasoning is generated under BAGEL's thinking system prompt, and a separate image-generation call is conditioned on that system prompt, the prompt, and the generated reasoning (text guidance 4, image guidance 1.5).

How often the fallback is used depends strongly on the prompt format. On our only surviving post-trained multimodal checkpoint, a Multimodal RWR variant whose reward additionally includes an OCR score on long-text prompts (not the Multimodal RWR checkpoint reported in the main tables), which is also used in Appendix~\ref{app:switching}, the model emitted the image-start token itself for 99.8\% of GenEval prompts, 88.2\% of DPG-Bench prompts, 40.7\% of OneIG-Bench prompts (none of its 225 Knowledge-Reasoning prompts), and 33.2\% of WISE prompts. For the remaining prompts it typically answered or described the prompt directly in text, and the image was produced by the manual modality switch. When emitted, the token always immediately followed \thinkendtok{}. These rates were measured on that checkpoint; the evaluation traces of the checkpoints reported in the main tables were not retained.

We generate four images per prompt for GenEval, four images per prompt arranged as a 2$\times$2 grid of 512$\times$512 images for DPG-Bench and OneIG-Bench, and one image per prompt for WISE.

\paragraph{Baselines.} The BAGEL Image-Only GenEval baseline uses the released EMA weights with text guidance 4, timestep shift 3, 50 steps, 1024$\times$1024 images, and seed 9775; the spread across 11 seeds is reported in Section~\ref{sec:modality_experiments_results}. The BAGEL DPG-Bench rows use the released EMA weights and 2$\times$2 grids of 512$\times$512 images. The BAGEL rows for WISE and OneIG-Bench are taken from prior work rather than regenerated with our pipeline.

\paragraph{Image-only post-trained models.} Image-Only SFT and Image-Only RWR generate images directly from the benchmark prompt without reasoning, using the same generation settings and weights as our other post-trained models (512$\times$512, 50 steps, timestep shift 1.0, text guidance 10, non-EMA weights). Their WISE images are scored only with GPT-6.1 Sol (Table~\ref{tab:wise_common_judge}), because the GPT-4o judge used for all other WISE results had become unavailable.

%% Switching-mode comparison (rebuttal)

\section{Learned versus Externally Controlled Modality Switching}
\label{app:switching}
To measure whether the learned modality switch affects generation quality, we compare three modality-switching methods on our only surviving post-trained multimodal checkpoint, a Multimodal RWR variant whose reward additionally includes an OCR score on long-text prompts (not the Multimodal RWR checkpoint reported in the main tables), on GenEval (553 prompts, 4 images each). For the first two methods we decode one greedy reasoning trace per prompt and reuse it, so they differ only in the image-generation context: (i) the \emph{learned modality switch}, our evaluation setting, in which the model emits \vstarttok{} and the image is conditioned on the generated reasoning; (ii) a \emph{\thinkendtok{}-triggered modality switch}, in which image generation is started at \thinkendtok{} and the image is conditioned on the prompt followed by the same reasoning, with the same generation settings. We also run (iii) a \emph{manual modality switch} using BAGEL's standard thinking pipeline on the same checkpoint (thinking system prompt, sampled reasoning, and a separate image-generation call conditioned on the system prompt, prompt, and reasoning, with BAGEL's default generation settings) at 1024$\times$1024 and at 512$\times$512.

\begin{table}[h]
\centering
\small
\caption{GenEval scores under different modality-switching methods, all on our surviving post-trained checkpoint, a Multimodal RWR variant trained with an additional OCR reward on long-text prompts (not the Multimodal RWR checkpoint reported in the main tables). Differences are tested with a paired bootstrap over prompts; all 95\% confidence intervals of pairwise differences include zero.}
\label{tab:switching_modes}
\begin{tabular}{llc}
\toprule
Modality switch & Image-generation context & GenEval Overall \\
\midrule
Learned (our evaluation) & generated reasoning & 0.801 \\
\thinkendtok{}-triggered & prompt + same reasoning & 0.810 \\
Manual (BAGEL thinking pipeline), 1024$\times$1024 & system prompt + prompt + sampled reasoning & 0.810 \\
Manual (BAGEL thinking pipeline), 512$\times$512 & system prompt + prompt + sampled reasoning & 0.797 \\
\bottomrule
\end{tabular}
\end{table}

All pairwise differences are within 0.013, comparable to the variation from regenerating the same method with different noise (0.001, 95\% CI $[-0.015, 0.013]$). The post-trained model emits the token on 552 of 553 GenEval prompts, always immediately after \thinkendtok{}; for the remaining prompt our evaluation falls back to the manual modality switch (scoring that prompt as a failure instead gives 0.799). On benchmarks whose prompts read as questions or long descriptions, the token is emitted much less often (Appendix~\ref{app:eval_protocol}), so this comparison characterizes switching when the token is emitted. Base BAGEL, given the same input, never emits the token. The generated reasoning restates every object in the prompt for 99.1\% of prompts, which explains why omitting the prompt from the image context has little effect here. These results support attributing the improvements we report to post-training rather than to the switching mechanism. GenEval prompts are short; on benchmarks with longer prompts the image context could matter more.

%% Prompt overlap analysis (rebuttal)

\section{Semantic Overlap Between Training Prompt Sets and Benchmarks}
\label{app:prompt_overlap}
We embed every prompt of the three training prompt sets and of the four evaluation benchmarks with the LLM2CLIP text encoder (the encoder OneIG-Bench uses for its Reasoning score), and for each benchmark prompt compute its maximum cosine similarity to each training set. As a reference we also compute the similarity between two random halves of each benchmark. Cosine magnitudes depend on prompt length and are comparable only within a row. A CLIP ViT-L/14 text encoder gives the same closest set in 16 of 22 benchmark categories.

\begin{table}[h]
\centering
\small
\caption{Mean maximum cosine similarity of benchmark prompts to each training prompt set, and the share of benchmark prompts whose nearest neighbor comes from each set (MMGW / GenEval-generated / Shutterstock).}
\label{tab:prompt_overlap}
\begin{tabular}{lccccc}
\toprule
Benchmark & Half-split reference & MMGW & GenEval-gen. & Shutterstock & Nearest-set share \\
\midrule
GenEval & 0.793 & 0.675 & 0.869 & 0.632 & 0 / 100 / 0\% \\
DPG-Bench & 0.621 & 0.324 & 0.280 & 0.368 & 24 / 12 / 64\% \\
WISE & 0.711 & 0.581 & 0.546 & 0.608 & 28 / 14 / 58\% \\
OneIG-Bench & 0.546 & 0.373 & 0.296 & 0.382 & 37 / 7 / 56\% \\
\quad OneIG Text & 0.504 & 0.426 & 0.257 & 0.334 & 82 / 2 / 16\% \\
\bottomrule
\end{tabular}
\end{table}

No training prompt matches a benchmark prompt after normalization. Near-duplicates (token Jaccard $\geq 0.8$) occur only between GenEval-generated prompts and GenEval (123 of 553 GenEval prompts), all of them template variants such as swapped relations or colors; 73\% of GenEval prompts have a GenEval-generated prompt with exactly the same objects. By simple keyword heuristics, 32\% of MMGW prompts request rendered text (0\% of GenEval-generated prompts, 4\% of Shutterstock captions), 23\% involve counting (4\% and 4\%), and 7\% use GenEval-style spatial relations (25\% and 1\%).

\clearpage
%% Training Hyperparameters
\section{Training Hyperparameters}
\label{app:training_hyperparams}

\begin{table}[h]
\centering
\setlength{\tabcolsep}{20pt} % increase space between columns
\renewcommand{\arraystretch}{2.5} % increase space between rows
\caption{Training hyperparameters used in all experiments in Sections \ref{sec:modality_experiments} and \ref{sec:data_strategies_experiments}.}
\begin{tabular}{l c}
\toprule
\textbf{Hyperparameter} & \textbf{Value} \\
\midrule
Total Training Steps & 5500 \\
Warm-Up Steps & \makecell{500 (1000 for CE-weighted RWR,\\ MSE-weighted RWR, and rejection sampling)} \\
Learning Rate (LR) & $5\times10^{-5}$ \\
LR scheduler & Constant \\
Sequence Length (per rank) & 50{,}000 (Tokens) \\
% Maximum number of tokens & 50{,}000 \\
Optimizer & AdamW $(\beta_1=0.9,\, \beta_2=0.95,\, \epsilon=1.0\times10^{-15})$ \\
Gradient Norm Clip Threshold & 1.0 \\
Loss Weights (CE, MSE) & (1.0, 0.1) \\
EMA Decay & 0.99 \\
RWR Temperature $\beta$ / Reward Normalization & 5.0 / global z-score, no clipping \\
\makecell[l]{Image Generation Resolution\\(min short side, max long side)} & (512, 512) \\
\makecell[l]{Image Understanding Resolution\\(min short side, max long side)} & (224, 518) \\
\bottomrule
\end{tabular}
\label{tab:training_hyperparams}
\end{table}

\clearpage

%% ImageReward Results
\section{ImageReward for RWR Results}
\label{app:imagereward_results}

Results using ImageReward~\cite{imagerewardlearningevaluatinghuman} as the reward weight during Reward-Weighted Regression post-training as described in Section~\ref{sec:rwr} are presented for GenEval~\cite{geneval} in Table~\ref{tab:geneval_modality_with_imagereward}, DPG-Bench~\cite{dpgbench} in Table~\ref{tab:dpgbench_modality_with_imagereward}, WISE~\cite{wiseworldknowledgeinformedsemantic} in Table~\ref{tab:wise_modality_with_imagereward}, and OneIG-Bench~\cite{oneigbenchomnidimensionalnuancedevaluation} in Table~\ref{tab:oneigbench_modality_with_imagereward}. 
Across all benchmarks, QwenVQAScore as the RWR reward weight outperforms ImageReward. 
Notably on WISE (GPT-4o judge), the knowledge-based image generation benchmark, all three modality weighting applications of ImageReward regress the overall score compared to the BAGEL Multimodal base model; this suggests that ImageReward is an especially low-quality reward for T2I tasks that require implicit understanding of world knowledge. 
These results are consistent with the reward validation in Section~\ref{sec:rewards}, where QwenVQAScore agrees best with independent judgments of generation success.

\input{tables/geneval_modality_with_imagereward}
\input{tables/dpg_modality_with_imagereward}
\input{tables/wise_modality_with_imagereward}
\input{tables/oneig_modality_with_imagereward}

\clearpage

%% Rejection Sampling
\section{Rejection Sampling Comparison}
\label{app:rejection_sampling}

As an alternative to reward-weighted regression, we also evaluated rejection sampling: fine-tuning only on samples with QwenVQAScore $\geq 0.5$ (the success threshold used in Section~\ref{sec:data_strategies_experiments}), with all retained samples weighted equally. Of the 357,300 samples in our MMGW training set, 204,181 (57.1\%) clear this threshold and are retained for rejection sampling training. Results across all four benchmarks are presented in Table~\ref{tab:rejection_samp}, alongside every post-training configuration from Section~\ref{sec:modality_experiments}.
Multimodal RWR outperforms rejection sampling on every benchmark and metric we report.
Notably, rejection sampling also fails to improve over the BAGEL Multimodal baseline on GenEval, DPG-Bench, WISE (GPT-4o judge), and the OneIG-Bench Text category, suggesting that the gains come from how the reward is applied during training rather than from filtering on reward alone.

\input{tables/rejection_sampling}

\clearpage

%% More Intra-Prompt Reward Distributions
\section{Reward Distributions}
\label{app:reward_dists}

\begin{figure}[H]
  \centering
  \includegraphics[width=0.8\textwidth]{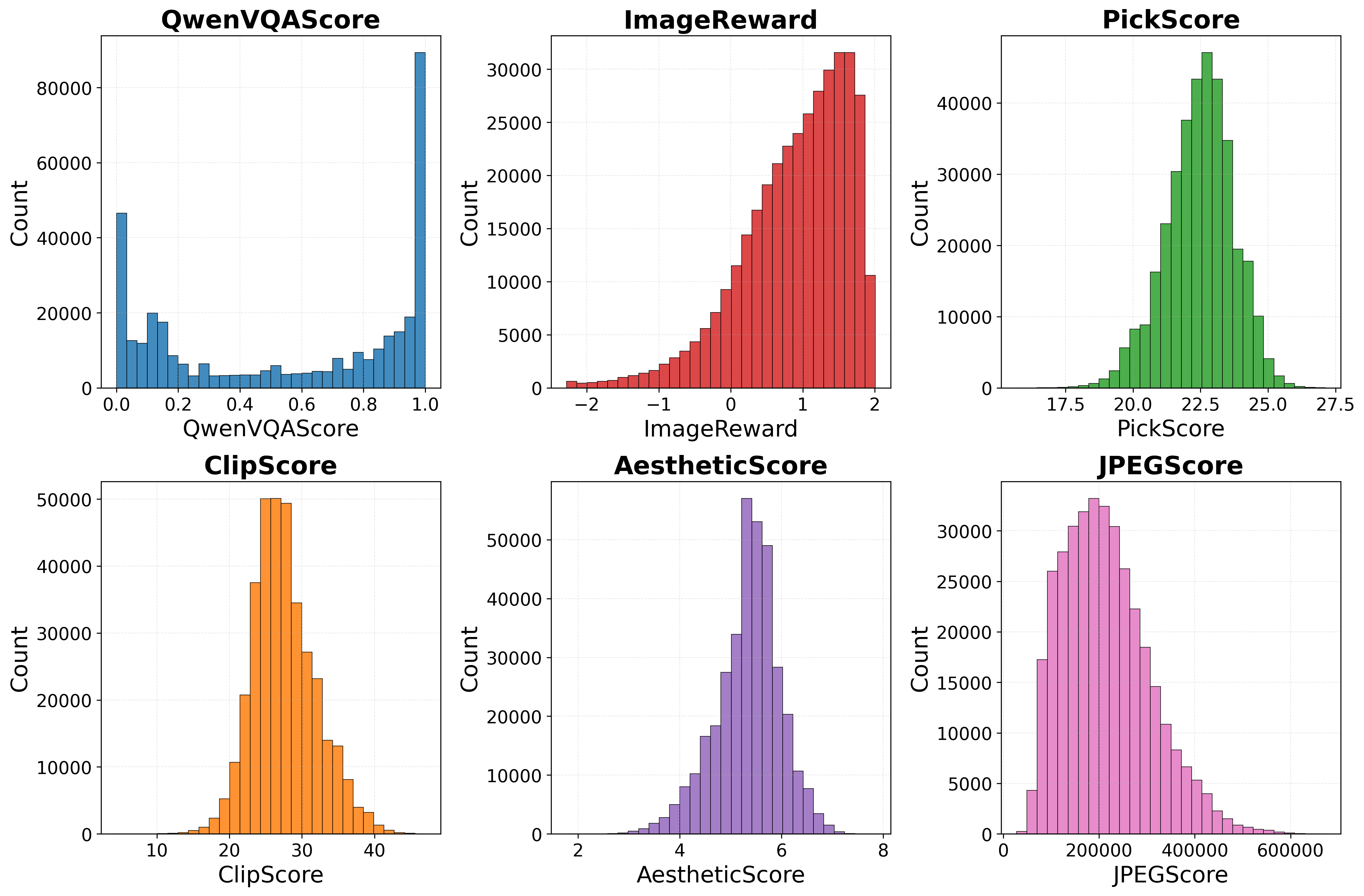}
  \caption{Reward distributions across all samples. QwenVQAScore exhibits a distinct bimodal distribution, while the other reward functions are unimodal. Distribution shape alone does not establish that a reward separates successful from failed generations; we validate each reward against independent references in Table~\ref{tab:reward_validation}.}
  \label{fig:global_reward_dist}
\end{figure}

\begin{figure}[H]
    \centering
    \includegraphics[width=1.0\textwidth]{figs/5_jars.pdf}
    
    \includegraphics[width=1.0\textwidth]{figs/plushie_reward.pdf}

    \includegraphics[width=1.0\textwidth]{figs/lighthouse_left.pdf}

    \caption{Reward distributions for individual prompts show that QwenVQAScore effectively scores generations and produces intra-prompt rewards that distinguish good from bad samples. Table~\ref{tab:reward_validation} compares all rewards quantitatively against independent judgments of good and bad samples, including human annotations.}
    \vspace{-0.5cm}
\end{figure}

%% file: tables/geneval_dataset.tex
\begin{table}[htbph]
% \vspace{-1.5em}
\centering
\small
\caption{
    \textbf{Results on the GenEval Benchmark}; The best score is in \colorbox{best_result_color}{green} and the second-best score in \colorbox{second_best_result_color}{orange}. 
}
\label{tab:geneval_datasets}
\resizebox{1.0\textwidth}{!}{
\begin{tabular}
{l@{\hspace{0.5cm}}ccccccc}
\toprule
\multicolumn{1}{c}
{\multirow{2}{*}{\bf \shortstack{Training Prompt\\Set Ablation} }} & \multirow{2}{*}{\bf \shortstack{Single\\Object}}  & \multirow{2}{*}{\bf \shortstack{Two\\Object}} & \multirow{2}{*}{\bf Counting} & \multirow{2}{*}{\bf Colors} & \multirow{2}{*}{\bf Position} & \multirow{2}{*}{\bf \shortstack{Color\\Attribute}} & {\multirow{2}{*}{\bf Overall $\uparrow$}}
\\
&&&&&&& \\
\cmidrule{1-8}

Shutterstock & 0.99 & 0.95 & 0.73 & 0.91 & 0.53 & 0.66 & 0.79 \\
GenEval-Generated~\cite{flowgrpotrainingflowmatching} & 0.98 & 0.95 & 0.81 & 0.90 & 0.62 & 0.63 & \colorbox{second_best_result_color}{0.82} \\
MMGW (Ours) & 0.99 & 0.97 & 0.82 & 0.88 & 0.61 & 0.72 & \colorbox{best_result_color}{0.83} \\

\cmidrule[0.05pt]{1-8}
BAGEL Multimodal (Baseline)~\cite{bagel} & 0.99 & 0.94 & 0.79 & 0.86 & 0.50 & 0.62 & 0.78 \\
BAGEL Image-Only (Baseline)~\cite{bagel} & 0.99 & 0.93 & 0.82 & 0.86 & 0.52 & 0.62 & 0.79 \\

\bottomrule
\end{tabular}
}
\end{table}

%% file: tables/dpg_dataset.tex
\begin{table}[htbph]
% \vspace{-1.5em}
\centering
\small
\caption{
    \textbf{Results on the DPG-Bench Benchmark}; The best score is in \colorbox{best_result_color}{green} and the second-best score in \colorbox{second_best_result_color}{orange}. 
}
\label{tab:dpgbench_datasets}
\resizebox{1.0\textwidth}{!}{
\begin{tabular}
{l@{\hspace{0.5cm}}cccccc}
\toprule
\multicolumn{1}{c}
{\multirow{2}{*}{\bf \shortstack{Training Prompt\\Set Ablation} }} & \multirow{2}{*}{\bf Global}  & \multirow{2}{*}{\bf Entity} & \multirow{2}{*}{\bf Attribute} & \multirow{2}{*}{\bf Relation} & \multirow{2}{*}{\bf Other} & {\multirow{2}{*}{\bf Overall $\uparrow$}}
\\
&&&&&& \\
\cmidrule{1-7}

Shutterstock & 91.60 & 86.81 & 88.23 & 91.26 & 91.87 & 84.23 \\
GenEval-Generated~\cite{flowgrpotrainingflowmatching} & 90.36 & 88.46 & 88.53 & 88.31 & 89.72 & 83.16 \\
MMGW (Ours) & 89.20 & 90.63 & 89.73 & 90.03 & 86.81 & \colorbox{second_best_result_color}{84.37} \\

\cmidrule[0.05pt]{1-7}
BAGEL Multimodal (Baseline)~\cite{bagel} & 88.75 & 90.84 & 89.01 & 89.44 & 89.70 & 84.01 \\
BAGEL Image-Only (Baseline)~\cite{bagel} & 90.73 & 90.05 & 89.72 & 92.23 & 90.80 & \colorbox{best_result_color}{85.08} \\

\bottomrule
\end{tabular}
}
\end{table}

%% file: tables/wise_datasets.tex
\begin{table}[htbph]
% \vspace{-1.5em}
\centering
\small
\caption{
    \textbf{Results on the WISE Benchmark \cite{wiseworldknowledgeinformedsemantic} (GPT-4o judge)}; All scores use the GPT-4o judge of the original WISE protocol; BAGEL rows are taken from \citet{bagel}. The best score is in \colorbox{best_result_color}{green} and the second-best score in \colorbox{second_best_result_color}{orange}. 
}
\label{tab:wise_datasets}
\resizebox{\textwidth}{!}{
\begin{tabular}
{l@{\hspace{0.5cm}}ccccccc}
\toprule
\multicolumn{1}{c}
{\multirow{2}{*}{\bf \shortstack{Training Prompt\\Set Ablation} }} & \multirow{2}{*}{\bf Cultural}  & \multicolumn{2}{c}{\bf Spatio-Temporal} & \multicolumn{3}{c}{\bf Natural Science } & {\multirow{2}{*}{\bf Overall $\uparrow$}}
\\
\cmidrule(lr){3-4}\cmidrule(lr){5-7}
&&
{\bf Time} &
{\bf Space} &
{\bf Biology  } &
{\bf Physics} &
{\bf Chemistry} &
\\
\cmidrule{1-8}

Shutterstock & 0.72 & 0.62 & 0.69 & 0.58 & 0.74 & 0.62 & 0.66 \\
GenEval-Generated~\cite{flowgrpotrainingflowmatching} & 0.71 & 0.68 & 0.74 & 0.63 & 0.76 & 0.66 & \colorbox{second_best_result_color}{0.70} \\
MMGW (Ours) & 0.76 & 0.69 & 0.77 & 0.64 & 0.77 & 0.66 & \colorbox{best_result_color}{0.72} \\

\cmidrule[0.05pt]{1-8}
BAGEL Multimodal (Baseline)~\cite{bagel} & 0.76 & 0.69 & 0.75 & 0.65 & 0.75 & 0.58 & \colorbox{second_best_result_color}{0.70} \\
BAGEL Image-Only (Baseline)~\cite{bagel} & 0.44 & 0.55 & 0.68 & 0.44 & 0.60 & 0.39 & 0.52 \\

\bottomrule
\end{tabular}
}
\end{table}

%% file: tables/oneig_datasets.tex
% \begin{table}[ht]
% \centering
\begin{center}
\small
\captionof{table}{
    \textbf{Results on the OneIG-Bench Benchmark}; The best score is in \colorbox{best_result_color}{green} and the second-best score in \colorbox{second_best_result_color}{orange}. 
}
\label{tab:oneig_datasets}
\resizebox{\textwidth}{!}{
\begin{tabular}
{l@{\hspace{0.5cm}}ccccc}
\toprule
{\bf \shortstack{Training Prompt\\Set Ablation} } & {\bf Alignment $\uparrow$}  & {\bf Text $\uparrow$} & {\bf Reasoning $\uparrow$} & {\bf Style $\uparrow$} & {\bf Diversity $\uparrow$}
\\
\cmidrule{1-6}

Shutterstock & \colorbox{second_best_result_color}{0.801} & 0.008 & \colorbox{second_best_result_color}{0.219} & 0.369 & 0.145 \\
GenEval-Generated~\cite{flowgrpotrainingflowmatching} & 0.797 & 0.006 & 0.217 & 0.358 & 0.147 \\
MMGW (Ours) & \colorbox{best_result_color}{0.802} & \colorbox{second_best_result_color}{0.189} & \colorbox{best_result_color}{0.220} & \colorbox{second_best_result_color}{0.370} & 0.141 \\

\cmidrule[0.05pt]{1-6}
BAGEL Multimodal (Baseline)~\cite{bagel} & 0.793 & 0.020 & 0.206 & \colorbox{best_result_color}{0.390} & \colorbox{second_best_result_color}{0.209} \\
BAGEL Image-Only (Baseline)~\cite{bagel} & 0.769 & \colorbox{best_result_color}{0.244} & 0.173 & 0.367 & \colorbox{best_result_color}{0.251} \\

\bottomrule
\vspace{-1.5em}
\end{tabular}
}
\end{center}
% \end{table}

%% file: tables/geneval_modality_with_imagereward.tex
\begin{table*}[h]
% \vspace{-1.5em}
\centering
\small 
\caption{
    \textbf{Results on the GenEval Benchmark}; The best score is in \colorbox{best_result_color}{green} and the second-best score in \colorbox{second_best_result_color}{orange}. \multimodalbox{Blue border} indicates the best Multimodal score and \imageonlybox{purple border} indicates the best Image-Only score. $\dagger$ denotes base model baseline and $*$ denotes SFT baseline, trained on the same data as RWR but without reward-weighting.
}
\vspace{-0.2cm}
\label{tab:geneval_modality_with_imagereward}
\begin{adjustbox}{max width=1.0\textwidth}
\begin{tabular}[h]
{l@{\hspace{0.5cm}}l@{\hspace{0.5cm}}ccccccc}
\toprule
{\bf \shortstack{Modality\\Ablation}} & {\bf Model} & {\bf \shortstack{Single\\Object}}  & {\bf \shortstack{Two\\Object}} & {\bf Counting} & {\bf Colors} & {\bf Position} & {\bf \shortstack{Color\\Attribute}} & {\bf Overall $\uparrow$}
\\
\midrule

\multirow{5}{*}{Image-Only} 
& BAGEL Image-Only $^\dagger$~\cite{bagel} & 0.99 & 0.93 & 0.82 & 0.86 & 0.52 & 0.62 & 0.79 \\
& Image-Only SFT$^*$ & 1.00 & 0.95 & 0.76 & 0.88 & 0.46 & 0.65 & 0.78 \\
& Image-Only RWR: QwenVQAScore & 1.00 & 0.93 & 0.81 & 0.88 & 0.51 & 0.68 & \imageonlyboxsecondbest{0.80} \\
& SD3-Medium~\cite{sd3_esser2024scalingrectifiedflowtransformers} & 0.99 & 0.94 & 0.72 & 0.89 & 0.33 & 0.60 & 0.74 \\
& FLUX.1[dev]~\cite{flux1kontextflowmatching} & 0.98 & 0.81 & 0.74 &  0.79 & 0.22 & 0.45 & 0.66 \\
\midrule

\multirow{7}{*}{Multimodal} 
& Show-o~\cite{show-o} & 0.98 & 0.80 & 0.66 & 0.84 & 0.31 & 0.50 & 0.68 \\
& Janus-Pro-7B~\cite{januspro} & 0.99 & 0.89 & 0.59 & 0.90 & 0.79 & 0.66 & \colorbox{second_best_result_color}{0.80} \\
& BAGEL Multimodal $^\dagger$~\cite{bagel} & 0.99 & 0.94 & 0.79 & 0.86 & 0.50 & 0.62 & 0.78 \\
& Multimodal SFT$^*$ & 0.99 & 0.93 & 0.79 & 0.86 & 0.54 & 0.64 & 0.79 \\
& CE-weighted RWR: QwenVQAScore & 1.00 & 0.93 & 0.79 & 0.88 & 0.52 & 0.60 & 0.79 \\
& CE-weighted RWR: ImageReward & 1.00 & 0.93 & 0.75 & 0.88 & 0.53 & 0.63 & 0.78 \\
& MSE-weighted RWR: QwenVQAScore & 0.99 & 0.95 & 0.81 & 0.91 & 0.52 & 0.63 & \colorbox{second_best_result_color}{0.80} \\
& MSE-weighted RWR: ImageReward & 0.98 & 0.93 & 0.64 & 0.88 & 0.53 & 0.66 & 0.77 \\
& Multimodal RWR: QwenVQAScore & 0.99 & 0.97 & 0.82 & 0.88 & 0.61 & 0.72 & \multimodalboxbest{0.83} \\
& Multimodal RWR: ImageReward & 1.00 & 0.93 & 0.77 & 0.87 & 0.58 & 0.67 & \colorbox{second_best_result_color}{0.80} \\

\bottomrule
\vspace{-0.5cm}
\end{tabular}
\end{adjustbox}
\end{table*}

%% file: tables/dpg_modality_with_imagereward.tex
\begin{table*}[h]
% \vspace{-1.5em}
\centering
\small 
\caption{
    \textbf{Results on the DPG-Bench Benchmark}; Highlighting and markers as in Table~\ref{tab:geneval_modality_with_imagereward}.
    % The best score is in \colorbox{best_result_color}{green} and the second-best score in \colorbox{second_best_result_color}{orange}. \multimodalbox{Blue border} indicates the best Multimodal score and \imageonlybox{purple border} indicates the best Image-Only score. $\dagger$ denotes base model baseline and $*$ denotes SFT baseline, trained on the same data as RWR but without reward-weighting.
}
\vspace{-0.2cm}
\label{tab:dpgbench_modality_with_imagereward}
\begin{adjustbox}{max width=1.0\textwidth}
\begin{tabular}
{ll@{\hspace{0.5cm}}cccccc}
\toprule
{\bf \shortstack{Modality\\Ablation} } & {\bf Model} & {\bf Global}  & {\bf Entity} & {\bf Attribute} & {\bf Relation} & {\bf Other} & {\bf Overall $\uparrow$}
\\
\midrule

\multirow{5}{*}{Image-Only} 
& BAGEL Image-Only$^{\dagger}$~\cite{bagel} & 90.73 & 90.05 & 89.72 & 92.23 & 90.80 & 85.08 \\
& Image-Only SFT$^{\ast}$ & 88.53 & 93.92 & 89.66 & 91.32 & 91.46 & \imageonlyboxbest{86.98} \\
& Image-Only RWR: QwenVQAScore & 90.07 & 92.62 & 90.03 & 92.63 & 91.84 & \colorbox{second_best_result_color}{86.75} \\
& FLUX.1[dev]~\cite{flux1kontextflowmatching} & 74.35 & 90.00 & 88.96 & 90.87 & 88.33 & 83.84 \\
& SD3-Medium~\cite{sd3_esser2024scalingrectifiedflowtransformers} & 87.90 & 91.01 &  88.83 & 80.70 & 88.68 & 84.08 \\
\cmidrule{1-8}

\multirow{7}{*}{Multimodal} 
& BLIP3-o~\cite{blip3o} & - & - & - & - & - & 81.60 \\
& Janus-Pro-7B~\cite{januspro}& 86.90 & 88.90 & 89.40 & 89.32 & 89.48 & 84.19 \\
& BAGEL Multimodal$^{\dagger}$~\cite{bagel} & 88.75 & 90.84 & 89.01 & 89.44 & 89.70 & 84.01 \\
& Multimodal SFT$^{\ast}$ & 88.73 & 88.92 & 88.89 & 90.04 & 87.02 & 83.57 \\
& CE-weighted RWR: QwenVQAScore & 89.94 & 89.10 & 89.24 & 89.41 & 90.58 & 83.84 \\
& CE-weighted RWR: ImageReward & 89.71 & 88.32 & 88.23 & 89.89 & 89.77 & 83.51 \\
& MSE-weighted RWR: QwenVQAScore & 90.39 & 90.87 & 88.74 & 87.67 & 86.97 & 83.60 \\
& MSE-weighted RWR: ImageReward & 85.33 & 89.70 & 88.75 & 89.17 & 91.22 & 83.73 \\
& Multimodal RWR: QwenVQAScore & 89.20 & 90.63 & 89.73 & 90.03 & 86.81 & \multimodalbox{84.37} \\
& Multimodal RWR: ImageReward & 80.61 & 88.30 & 88.74 & 91.28 & 87.34 & 83.18 \\

\bottomrule
\vspace{-0.7cm}
\end{tabular}
\end{adjustbox}
\end{table*}

%% file: tables/wise_modality_with_imagereward.tex
\begin{table*}[]
% \vspace{-1.5em}
\centering
\small
\caption{
    \textbf{Results on the WISE Benchmark \cite{wiseworldknowledgeinformedsemantic} (GPT-4o judge)}; All scores use the GPT-4o judge of the original WISE protocol; rows for BAGEL and prior models are taken from their papers. 
    The best score is in \colorbox{best_result_color}{green} and the second-best score in \colorbox{second_best_result_color}{orange}. \multimodalbox{Blue border} indicates the best Multimodal score and \imageonlybox{purple border} indicates the best Image-Only score. $\dagger$ denotes base model baseline and $*$ denotes SFT baseline, trained on the same data as RWR but without reward-weighting.
 $^{\S}$Not scored with the GPT-4o judge, which became unavailable; GPT-6.1 Sol scores, not comparable to these, are in Table~\ref{tab:wise_common_judge}.
}
\vspace{-0.2cm}
\label{tab:wise_modality_with_imagereward}
\begin{adjustbox}{max width=1.0\textwidth}
\begin{tabular}
{l@{\hspace{0.5cm}}l@{\hspace{0.5cm}}ccccccc}
\toprule
\multirow{2}{*}{\bf \shortstack{Modality\\Ablation}} & \multirow{2}{*}{\bf Model}  & \multirow{2}{*}{\bf Cultural}  & \multicolumn{2}{c}{\bf Spatio-Temporal} & \multicolumn{3}{c}{\bf Natural Science } & {\multirow{2}{*}{\bf Overall $\uparrow$}}
\\
\cmidrule(lr){4-5}\cmidrule(lr){6-8}
&&&
{\bf Time} &
{\bf Space} &
{\bf Biology  } &
{\bf Physics} &
{\bf Chemistry} &
\\
\cmidrule{1-9}

% Image-Only models
\multirow{6}{*}{Image-Only} 
& BAGEL Image-Only$^\dagger$~\cite{bagel} & 0.44 & 0.55 & 0.68 & 0.44 & 0.60 & 0.39 & \imageonlybox{0.52} \\
& Image-Only SFT$^*$ & -- & -- & -- & -- & -- & -- & --$^{\S}$ \\
& Image-Only RWR: QwenVQAScore & -- & -- & -- & -- & -- & -- & --$^{\S}$ \\
& FLUX.1[dev]~\cite{flux1kontextflowmatching} & 0.48 & 0.58 & 0.62 & 0.42 & 0.51 & 0.35 & 0.50 \\
& SD3.5 Large~\cite{sd3_esser2024scalingrectifiedflowtransformers} & 0.44 & 0.50 & 0.58 & 0.44 & 0.52 & 0.31 & 0.46 \\
& SD XL~\cite{sdxlimprovinglatentdiffusion} & 0.43 & 0.48 & 0.47 & 0.44 & 0.45 & 0.27 & 0.43 \\
\cmidrule{1-9}

% Multimodal models
\multirow{9}{*}{Multimodal} 
& BLIP3-o~\cite{blip3o} & - & - & - & - & - & - & 0.62 \\
& T2I-R1~\cite{t2ir1reinforcingimagegeneration} & 0.56 & 0.55 & 0.63 & 0.54 & 0.55 & 0.30 & 0.54 \\
& Show-o~\cite{show-o} & 0.28 & 0.40 & 0.48 & 0.30 & 0.46 & 0.30 & 0.35 \\
& Janus-Pro-7B~\cite{januspro}& 0.30 & 0.37 & 0.49 & 0.36 & 0.42 & 0.26 & 0.35 \\
& BAGEL Multimodal$^\dagger$~\cite{bagel} & 0.76 & 0.69 & 0.75 & 0.65 & 0.75 & 0.58 & \colorbox{second_best_result_color}{0.70} \\
& Multimodal SFT$^*$ & 0.70 & 0.64 & 0.72 & 0.60 & 0.77 & 0.65 & 0.68 \\
& CE-weighted RWR: QwenVQAScore & 0.73 & 0.65 & 0.75 & 0.60 & 0.80 & 0.63 & 0.69 \\
& CE-weighted RWR: ImageReward & 0.71 & 0.62 & 0.75 & 0.58 & 0.74 & 0.60 & 0.67 \\
& MSE-weighted RWR: QwenVQAScore & 0.73 & 0.67 & 0.74 & 0.64 & 0.80 & 0.65 & \colorbox{second_best_result_color}{0.70} \\
& MSE-weighted RWR: ImageReward & 0.72 & 0.63 & 0.72 & 0.59 & 0.74 & 0.62 & 0.67 \\
& Multimodal RWR: QwenVQAScore & 0.76 & 0.69 & 0.77 & 0.64 & 0.77 & 0.66 & \multimodalboxbest{0.72} \\
& Multimodal RWR: ImageReward & 0.72 & 0.65 & 0.74 & 0.59 & 0.76 & 0.63 & 0.68 \\

\bottomrule
\vspace{-0.8cm}
\end{tabular}
\end{adjustbox}
\end{table*}

%% file: tables/oneig_modality_with_imagereward.tex
\begin{table*}[]
% \vspace{-1.5em}
\centering
\small
\caption{
    \textbf{Results on the OneIG-Bench Benchmark}; Highlighting and markers as in Table~\ref{tab:geneval_modality_with_imagereward}.
    % The best score is in \colorbox{best_result_color}{green} and the second-best score in \colorbox{second_best_result_color}{orange}. $\dagger$ denotes base model baseline. $*$ denotes SFT baseline, trained on the same data as RWR but without reward-weighting.
}
\label{tab:oneigbench_modality_with_imagereward}
\begin{adjustbox}{max width=1.0\textwidth}
\begin{tabular}
{l@{\hspace{0.5cm}}l@{\hspace{0.5cm}}ccccc}
\toprule
{\bf \shortstack{Modality\\Ablation}} & {\bf Model} & {\bf Alignment $\uparrow$}  & {\bf Text $\uparrow$} & {\bf Reasoning $\uparrow$} & {\bf Style $\uparrow$} & {\bf Diversity $\uparrow$}
\\
\midrule

\multirow{6}{*}{Image-Only}
& BAGEL Image-Only$^\dagger$~\cite{bagel} & 0.769 & 0.244 & 0.173 & 0.367 & 0.251 \\
& Image-Only SFT$^*$ & \imageonlyboxbest{0.837} & 0.444 & 0.209 & 0.374 & 0.179 \\
& Image-Only RWR: QwenVQAScore & \imageonlyboxbest{0.837} & 0.475 & 0.210 & \imageonlyboxsecondbest{0.377} & 0.184 \\
& FLUX.1[dev]~\cite{flux1kontextflowmatching} & 0.786 &  \colorbox{second_best_result_color}{0.523} & \colorbox{second_best_result_color}{0.253} & 0.368 & 0.238 \\
& SD XL~\cite{sdxlimprovinglatentdiffusion} & 0.688 & 0.029 & 0.237 & 0.332 & \colorbox{second_best_result_color}{0.296} \\
% SD3.5 Large is #1 in Alignment, Text, Reasoning !!!
& SD3.5 Large~\cite{sd3_esser2024scalingrectifiedflowtransformers} & 
\colorbox{second_best_result_color}{0.809} & \imageonlyboxbest{0.629} & \imageonlyboxbest{0.294} & 0.353 & 0.225 \\
\cmidrule{1-7}

\multirow{8}{*}{Multimodal} 
& Show-o~\cite{show-o} & 0.702 & 0.002 & 0.213 & 0.361 & 0.241 \\
& BLIP3-o~\cite{blip3o} & 0.711 & 0.013 & \multimodalbox{0.223} & 0.361 & 0.229 \\
& Janus-Pro-7B~\cite{januspro}& 0.553 & 0.001 & 0.139 & 0.276 & \multimodalboxbest{0.365} \\
& BAGEL Multimodal$^\dagger$~\cite{bagel} & 0.793 & 0.020 & 0.206 & \multimodalboxbest{0.390} & 0.209 \\
& Multimodal SFT$^*$ & 0.800 & 0.010 & 0.219 & 0.365 & 0.143 \\
& CE-weighted RWR: QwenVQAScore & 0.803 & 0.057 & 0.220 & 0.366 & 0.149 \\
& CE-weighted RWR: ImageReward & 0.803 & 0.016 & 0.220 & 0.359 & 0.149 \\
& MSE-weighted RWR: QwenVQAScore & \multimodalbox{0.804} & 0.077 & 0.219 & 0.364 & 0.144 \\
& MSE-weighted RWR: ImageReward & 0.803 & 0.038 & 0.217 & 0.357 & 0.143 \\
& Multimodal RWR: QwenVQAScore & 0.802 & \multimodalbox{0.189} & 0.220 & 0.370 & 0.141 \\
& Multimodal RWR: ImageReward & 0.802 & 0.040 & 0.217 & 0.348 & 0.143 \\

\bottomrule
\vspace{-0.7cm}
\end{tabular}
\end{adjustbox}
\end{table*}

%% file: tables/rejection_sampling.tex
\begin{table}[h]
\centering
\small
\caption{
    \textbf{Rejection sampling compared to all post-training configurations.} Among post-training methods, the best score is in \colorbox{best_result_color}{green} and the second-best score in \colorbox{second_best_result_color}{orange}; baseline rows are shown for reference and are excluded from highlighting. Image-Only SFT, Image-Only RWR, and BAGEL Image-Only generate images without an intermediate reasoning trace. WISE scores use the GPT-4o judge. $^{\S}$Not scored with the GPT-4o judge, which became unavailable; GPT-6.1 Sol scores, not comparable to these, are in Table~\ref{tab:wise_common_judge}.
}
\label{tab:rejection_samp}
\resizebox{\textwidth}{!}{
\begin{tabular}{l@{\hspace{0.3cm}}c@{\hspace{0.3cm}}c@{\hspace{0.3cm}}c@{\hspace{0.3cm}}ccccc}
\toprule
\multirow{2}{*}{\bf \shortstack{Post-Training\\Method}} &
\multirow{2}{*}{\bf \shortstack{GenEval\\Overall $\uparrow$}} &
\multirow{2}{*}{\bf \shortstack{DPG-Bench\\Overall $\uparrow$}} &
\multirow{2}{*}{\bf \shortstack{WISE (GPT-4o)\\Overall $\uparrow$}} &
\multicolumn{5}{c}{\bf OneIG-Bench}
\\
\cmidrule(lr){5-9}
& & & &
{\bf Alignment $\uparrow$} &
{\bf Text $\uparrow$} &
{\bf Reasoning $\uparrow$} &
{\bf Style $\uparrow$} &
{\bf Diversity $\uparrow$}
\\
\midrule

Image-Only SFT & 0.78 & \colorbox{best_result_color}{86.98} & --$^{\S}$ & \colorbox{best_result_color}{0.837} & \colorbox{second_best_result_color}{0.444} & 0.209 & \colorbox{second_best_result_color}{0.374} & \colorbox{second_best_result_color}{0.179} \\
Image-Only RWR & \colorbox{second_best_result_color}{0.80} & \colorbox{second_best_result_color}{86.75} & --$^{\S}$ & \colorbox{best_result_color}{0.837} & \colorbox{best_result_color}{0.475} & 0.210 & \colorbox{best_result_color}{0.377} & \colorbox{best_result_color}{0.184} \\
Multimodal SFT & 0.79 & 83.57 & 0.68 & 0.800 & 0.010 & 0.219 & 0.365 & 0.143 \\
CE-weighted RWR & 0.79 & 83.84 & 0.69 & 0.803 & 0.057 & \colorbox{best_result_color}{0.220} & 0.366 & 0.149 \\
MSE-weighted RWR & \colorbox{second_best_result_color}{0.80} & 83.60 & \colorbox{second_best_result_color}{0.70} & \colorbox{second_best_result_color}{0.804} & 0.077 & 0.219 & 0.364 & 0.144 \\
Multimodal RWR & \colorbox{best_result_color}{0.83} & 84.37 & \colorbox{best_result_color}{0.72} & 0.802 & 0.189 & \colorbox{best_result_color}{0.220} & 0.370 & 0.141 \\
Rejection Sampling & 0.78 & 83.82 & \colorbox{second_best_result_color}{0.70} & 0.801 & 0.009 & 0.219 & 0.359 & 0.134 \\

\midrule
BAGEL Multimodal (Baseline)~\cite{bagel} & 0.78 & 84.01 & 0.70 & 0.793 & 0.020 & 0.206 & 0.390 & 0.209 \\
BAGEL Image-Only (Baseline)~\cite{bagel} & 0.79 & 85.08 & 0.52 & 0.769 & 0.244 & 0.173 & 0.367 & 0.251 \\

\bottomrule
\end{tabular}
}
\end{table}